%% file: main.tex
\documentclass[final]{cvpr}

\usepackage{times}
\usepackage{epsfig}
\usepackage{graphicx}
\usepackage{amsmath}
\usepackage{amssymb}
\usepackage[all=normal,leading=tight]{savetrees} 

\usepackage{booktabs}            
\usepackage{comment}             
\usepackage{lscape}              
\usepackage{multirow}            
\newcommand*\rot{\rotatebox{90}} 

\usepackage[pagebackref=true,breaklinks=true,colorlinks,bookmarks=false]{hyperref}

\usepackage{subfiles} 

\DeclareMathOperator*{\argmin}{arg\,min}

\newcommand{\keypoint}[1]{\noindent\textbf{#1}\quad}
\newcommand{\cut}[1]{}

\def\cvprPaperID{7087} 
\def\confYear{CVPR 2021}

\usepackage{cuted}
\usepackage{capt-of}

\begin{document}

\title{How Well Do Self-Supervised Models Transfer?}

\author{Linus Ericsson\\
University of Edinburgh\\
~\\
{\tt\small linus.ericsson@ed.ac.uk}
\and
Henry Gouk\\
University of Edinburgh\\
~\\
{\tt\small henry.gouk@ed.ac.uk}
\and
Timothy M. Hospedales\\
University of Edinburgh, \\
Samsung AI Research, Cambridge\\
{\tt\small t.hospedales@ed.ac.uk}
}

\maketitle

\begin{strip}
    \centering
    \includegraphics[width=0.90\linewidth]{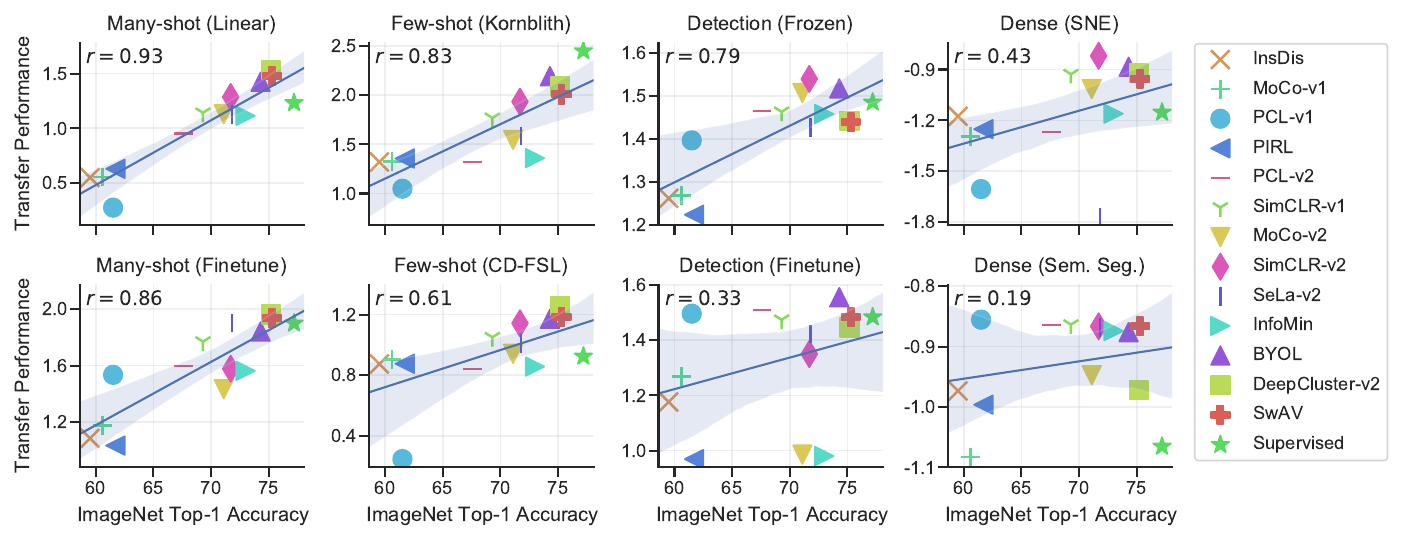}
    \captionof{figure}{Transfer performance is highly correlated with ImageNet performance for many-shot recognition but increasingly less correlated for few-shot recognition, object detection and dense prediction. On the x-axes we plot ImageNet top-1 accuracy and on the y-axes the average transfer log-odds. The gradients of the regression lines describe the correlation, with confidence intervals in shaded areas. For perfect correlation, the ideal line is a positive slope diagonal. Correlation coefficients (Pearson's $r$) are shown in the top left of each plot.}
    \label{fig:transfer_full}
\end{strip}

\begin{abstract}
   Self-supervised visual representation learning has seen huge progress recently, but no large scale evaluation has compared the many models now available. We evaluate the transfer performance of 13 top self-supervised models on 40 downstream tasks, including many-shot and few-shot recognition, object detection, and dense prediction. We compare their performance to a supervised baseline and show that on most tasks the best self-supervised models outperform supervision, confirming the recently observed trend in the literature. We find ImageNet Top-1 accuracy to be highly correlated with transfer to many-shot recognition, but increasingly less so for few-shot, object detection and dense prediction. No single self-supervised method dominates overall, suggesting that universal pre-training is still unsolved. Our analysis of features suggests that top self-supervised learners fail to preserve colour information as well as supervised  alternatives, but tend to induce better classifier calibration, and less attentive overfitting than supervised learners.
\end{abstract}

\subfile{sections/1-introduction}
\subfile{sections/2-related-work}
\subfile{sections/3-preliminaries}
\subfile{sections/4-experiments}
\subfile{sections/5-discussion}

\vspace{0.2cm}\keypoint{Acknowledgements} 
This research was partially supported by the Engineering and Physical Sciences Research Council (EPSRC) Grant number EP/S000631/1 and the MOD University Defence Research Collaboration (UDRC) in Signal Processing; EPSRC Centre for Doctoral Training in Data Science, funded by
EPSRC (grant EP/L016427/1) and the University
of Edinburgh; and EPSRC grant EP/R026173/1.

{\small
\bibliographystyle{ieee_fullname}
\bibliography{references}
}

\clearpage
\subfile{sections/7-appendix}

\end{document}

%% file: sections/1-introduction.tex
\section{Introduction}
\vspace{-0.1cm}
Computer vision in the last decade has been driven by increasingly sophisticated convolutional neural networks (CNNs) and the increasingly large datasets used to train them. Nevertheless, progress in this paradigm is ultimately bottlenecked by the data annotation process. This has motivated a growing wave of research in self-supervised representation learning, where CNN representations are trained on pretext tasks with freely available labels. Once trained, these CNN representations can be used to learn new tasks more data efficiently through feature re-use or finetuning.

Self-supervised learning (SSL) has been around for some time \cite{Schmidhuber1990MakingEnvironments}, but historically has lagged behind state of the art supervised representation learning. However, the recent pace of progress has increased dramatically and led to self-supervised deep representations that appear to approach and possibly even surpass that of fully-supervised representations \cite{Grill2020BootstrapLearning,Caron2020UnsupervisedAssignments}. This has raised hopes that self-supervised methods could indeed replace the ubiquitous annotation-intensive paradigm of supervised deep learning in state of the art computer vision going forward.

Given the growing practical importance of self-supervised learning as it approaches state of the art in computer vision tasks, there is increasing interest in understanding and benchmarking its empirical performance. Major recent evaluation studies have looked at aspects such as the fit between CNN architectures and choice of pretext task \cite{Kolesnikov2019RevisitingLearning} and the impact of the pre-training set size and CNN capacity on downstream task performance \cite{Goyal2019ScalingLearning}.

Despite this initial progress, there are a number of important open questions that remain to to be understood. Firstly, given the plethora of self-supervised representations on the market using diverse pre-text tasks and data-augmentations: which methods are the most empirically effective? This is currently hard to assess given the limited commonality in the evaluation conditions reported by each method. Secondly: While the most widely adopted benchmark metric is image classification performance, there are hopes that pre-trained representations will generalise to other downstream tasks such as detection and dense prediction \cite{Goyal2019ScalingLearning}. However, the published self-supervision literature is particularly inconsistent with regard to benchmarking these alternative tasks, making it impossible to determine the most effective methods. In particular, while we hope that the methods with best performance on the most popular benchmark of ImageNet recognition will also perform well on alternative tasks, this conjecture has never been systematically tested empirically. Thirdly: While core academic vision research is happy to focus on ImageNet as a benchmark, the wider community of computer vision practitioners work with diverse data types from medical \cite{Tschandl2018DataLesions} to agricultural \cite{Mohanty2016UsingDetection}, to earth-observation \cite{Helber2019Eurosat:Classification} data and beyond. From this perspective a crucial question is to what extent self-supervised features pre-trained on ImageNet can generalise directly to these diverse downstream tasks? This is important to know practically, because it dictates whether users in different vision domains can use pre-trained features directly, or whether they would need to collect their own datasets and perform domain-specific self-supervised learning -- a major data, compute and environmental \cite{schwartz2019greenAI} hurdle given that state of the art methods can take around 20 GPU days to train \cite{Chen2020ImprovedLearning}. Academically, this is also important to know, as an indicator of whether pursuing higher ImageNet accuracy in self-supervised learning research leads to higher accuracy on diverse real-world vision tasks, or is our research overfitting to ImageNet recognition?

To answer these questions and more, we conduct a large empirical benchmarking study on the efficacy of different pre-trained representations for diverse downstream tasks. In particular, we evaluate 13 pre-trained self-supervised models on 40 transfer tasks covering many-shot and few-shot image classification, object detection, surface normal prediction and semantic segmentation, as summarised in Fig.~\ref{fig:transfer_full}. Our downstream tasks cover diverse datasets with a wide range of similarity to the source ImageNet data, which all our models were pre-trained on.

Among other questions, we aim to answer the following:

\noindent\textbf{Q1.} \emph{How do state of the art self-supervised methods compare to supervised feature learning for diverse downstream datasets and tasks?} A: The best self-supervised methods can match and outperform supervised representation learning across most tasks considered. Only in few-shot recognition with small domain shift to ImageNet does supervised representation learning win.

\noindent\textbf{Q2.} \emph{Do self-supervised representations that perform well on ImageNet classification systematically perform well on diverse downstream datasets and tasks?} A: For recognition on datasets similar to ImageNet, performance is highly correlated. However, for some of the least similar recognition datasets such as ISIC2018, there is little to no correlation with ImageNet performance. For different tasks such as detection and dense prediction, correlation exists but is lower than for recognition.

\noindent\textbf{Q3.} \emph{Is there a best self-supervised representation overall?} A: No. For example, the recent methods SwAV and DeepCluster-v2 work well for recognition on ImageNet-like data, but under-perform on non-recognition tasks and on different data such as medical skin images. This suggests that the vision of a universal pre-trained model suited for all downstream tasks is yet to be realised.

\noindent\textbf{Q4.} \emph{Do self-supervised and supervised features represent the same information?} A: Contemporary self-supervised features seem to discard colour information, presumably due to the data augmentation they use. They also tend to be more attentively diffuse in contrast to the high spatial focus of attention in supervised features, which may contribute to their improved uncertainty calibration.

%% file: sections/2-related-work.tex
\section{Related Work}

\begin{table*}[t]
    \centering
    \caption{Top self-supervised models beat the supervised pre-training baseline on popular many-shot recognition datasets, both in linear evaluation and when finetuning. The top half of the table shows results from linear transfer of pre-trained models using logistic regression, and the bottom half shows the results when these models are finetuned. We also include the ImageNet linear evaluation performance (logistic regression or SGD) reported by the authors. Results style: \textbf{best}, \underline{second best}.}
    \label{tab:many_shot_evaluation}
    \resizebox{0.9\textwidth}{!}{%
    
    \begin{tabular}{clc|ccccccccccc|c}
    \toprule
    {} & {} &           ImageNet &           Aircraft &         Caltech101 &               Cars &            CIFAR10 &           CIFAR100 &                DTD &            Flowers &               Food &               Pets &             SUN397 &            VOC2007 &               Avg. \\
    \midrule
    \multirow{14}{*}{\rot{Linear}} & InsDis         &              59.50 &              36.87 &              71.12 &              28.98 &              80.28 &              59.97 &              68.46 &              83.44 &              63.39 &              68.78 &              49.47 &              74.37 &              62.29 \\
    {} & MoCo-v1        &              60.60 &              35.55 &              75.33 &              27.99 &              80.16 &              57.71 &              68.83 &              82.10 &              62.10 &              69.84 &              51.02 &              75.93 &              62.41 \\
    {} & PCL-v1         &              61.50 &              21.61 &              76.90 &              12.93 &              81.84 &              55.74 &              62.87 &              64.73 &              48.02 &              75.34 &              45.70 &              78.31 &              56.73 \\
    {} & PIRL           &              61.70 &              37.08 &              74.48 &              28.72 &              82.53 &              61.26 &              68.99 &              83.60 &              64.65 &              71.36 &              53.89 &              76.61 &              63.92 \\
    {} & PCL-v2         &              67.60 &              37.03 &              86.42 &              30.51 &              91.91 &              73.54 &              70.59 &              85.34 &              64.88 &              82.79 &              56.25 &              81.14 &              69.13 \\
    {} & SimCLR-v1      &              69.30 &              44.90 &              90.05 &              43.73 &              91.18 &              72.73 &              74.20 &              90.87 &              67.47 &              83.33 &              59.21 &              80.77 &              72.59 \\
    {} & MoCo-v2        &              71.10 &              41.79 &              87.92 &              39.31 &              92.28 &              74.90 &              73.88 &              90.07 &              68.95 &              83.30 &              60.32 &              82.69 &              72.31 \\
    {} & SimCLR-v2      &              71.70 &              46.38 &              89.63 &              50.37 &              92.53 &              76.78 &              76.38 &              92.90 &              73.08 &              84.72 &              61.47 &              81.57 &              75.07 \\
    {} & SeLa-v2        &              71.80 &              37.29 &              87.20 &              36.86 &              92.73 &              74.81 &              74.15 &              90.22 &              71.08 &              83.22 &              62.71 &              82.73 &              72.09 \\
    {} & InfoMin        &              73.00 &              38.58 &              87.84 &              41.04 &              91.49 &              73.43 &              74.73 &              87.18 &              69.53 &              86.24 &              61.00 &              83.24 &              72.21 \\
    {} & BYOL           &              74.30 &              53.87 &     \textbf{91.46} &  \underline{56.40} &              93.26 &              77.86 &              76.91 &              94.50 &              73.01 &              89.10 &              59.99 &              81.14 &              77.05 \\
    {} & DeepCluster-v2 &              75.20 &     \textbf{54.49} &  \underline{91.33} &     \textbf{58.60} &     \textbf{94.02} &     \textbf{79.61} &     \textbf{78.62} &     \textbf{94.72} &     \textbf{77.94} &  \underline{89.36} &  \underline{65.48} &     \textbf{83.94} &     \textbf{78.92} \\
    {} & SwAV           &  \underline{75.30} &  \underline{54.04} &              90.84 &              54.06 &  \underline{93.99} &  \underline{79.58} &  \underline{77.02} &  \underline{94.62} &  \underline{76.62} &              87.60 &     \textbf{65.58} &  \underline{83.68} &  \underline{77.97} \\
    \cmidrule{2-15}
    {} & Supervised     &     \textbf{77.20} &              43.59 &              90.18 &              44.92 &              91.42 &              73.90 &              72.23 &              89.93 &              69.49 &     \textbf{91.45} &              60.49 &              83.60 &              73.75 \\

    \midrule \toprule

    \multirow{14}{*}{\rot{Finetune}} & InsDis         & {} &              73.38 &              72.04 &              61.56 &              93.32 &              68.26 &              63.99 &              89.51 &              76.78 &              76.22 &              51.84 &              71.90 &              72.62 \\
    {} & MoCo-v1        & {} &              75.61 &              74.95 &              65.02 &              93.89 &              71.52 &              65.37 &              89.45 &              77.28 &              76.96 &              53.35 &              74.91 &              74.39 \\
    {} & PCL-v1         & {} &              74.97 &              87.62 &              73.24 &              96.35 &              79.62 &              70.00 &              90.83 &              78.30 &              86.98 &              58.40 &              82.08 &              79.85 \\
    {} & PIRL           & {} &              72.68 &              70.83 &              61.02 &              92.23 &              66.48 &              64.26 &              89.81 &              74.96 &              76.26 &              50.38 &              69.90 &              71.71 \\
    {} & PCL-v2         & {} &              79.37 &              88.04 &              71.68 &              96.50 &              80.26 &              71.76 &              92.95 &              80.34 &              85.39 &              58.82 &              82.20 &              80.66 \\
    {} & SimCLR-v1      & {} &              81.06 &              90.35 &              83.78 &     \textbf{97.07} &  \underline{84.53} &              71.54 &              93.75 &              82.40 &              84.10 &              63.31 &              82.58 &              83.13 \\
    {} & MoCo-v2        & {} &              79.87 &              84.38 &              75.20 &              96.45 &              71.33 &              69.47 &              94.35 &              76.78 &              79.80 &              55.77 &              71.71 &              77.74 \\
    {} & SimCLR-v2      & {} &              78.71 &              82.94 &              79.84 &              96.22 &              79.05 &              70.16 &              94.32 &              82.22 &              83.20 &              61.12 &              78.19 &              80.54 \\
    {} & SeLa-v2        & {} &              81.99 &              88.99 &              85.62 &              96.80 &              84.37 &              74.36 &     \textbf{95.80} &               86.24 &              88.55 &              65.84 &  \underline{84.85} &              84.86 \\
    {} & InfoMin        & {} &              80.24 &              83.92 &              78.76 &              96.94 &              71.15 &              71.12 &              95.24 &              78.93 &              85.28 &              57.66 &              76.63 &              79.62 \\
    {} & BYOL           & {} &              79.45 &              89.40 &              84.60 &              97.01 &              83.95 &              73.62 &              94.48 &              85.54 &  \underline{89.62} &              63.96 &              82.70 &              84.03 \\
    {} & DeepCluster-v2 & {} &              82.52 &  \underline{90.75} &     \textbf{87.27} &  \underline{97.06} &     \textbf{85.15} &  \underline{74.84} &              95.31 &     \textbf{87.51} &              89.43 &     \textbf{66.42} &     \textbf{84.90} &     \textbf{85.56} \\
    {} & SwAV           & {} &  \underline{83.08} &              89.85 &  \underline{86.76} &              96.78 &              84.37 &     \textbf{75.16} &              95.46 &  \underline{87.22} &              89.05 &  \underline{66.24} &              84.66 &  \underline{85.33} \\
    \cmidrule{2-15}
    {} & Supervised     & {} &     \textbf{83.50} &     \textbf{91.01} &              82.61 &              96.39 &              82.91 &              73.30 &  \underline{95.50} &              84.60 &     \textbf{92.42} &              63.56 &              84.76 &              84.60 \\
    \bottomrule

    \end{tabular}

    }

\end{table*}

\keypoint{Self-supervised learning} Self-supervised representation learning is now a large topic that it is impossible to cover completely here, and we point the reader to excellent recent surveys \cite{Jing2020Self-supervisedSurvey,Liu2020Self-supervisedContrastive} for thorough reviews. In this paper, we focus on still-image self-supervised learning, where a common paradigm is to pre-train on ImageNet \cite{Deng2009ImageNet:Database} using a variety of pre-text tasks from jigsaw puzzles \cite{Noroozi2016UnsupervisedPuzzles} to colorization \cite{Zhang2016ColorfulColorization,Larsson2016LearningColorization} to instance discrimination \cite{Wu2018UnsupervisedDiscrimination,Dosovitskiy2014DiscriminativeNetworks,Chen2020ARepresentations,He2019MomentumLearning} and clustering \cite{Li2020PrototypicalRepresentations,Caron2020UnsupervisedAssignments}. Evaluation is then typically performed by using the learned representation to train a linear classifier on ImageNet \cite{He2019MomentumLearning}, or finetune the representation with a small amount of data \cite{Chen2020BigLearners}. However, evaluation of the impact on different downstream datasets (where there is domain shift \cite{Zhang2020ImpactAdaptation} with respect to ImageNet), and non-recognition tasks has been highly inconsistent -- a gap in the literature that we aim to remedy in this paper.

To do this we wish to evaluate a large number of self-supervised methods, covering a wide range of training objectives. Many recent works adopt a form of instance discrimination \cite{Dosovitskiy2014DiscriminativeNetworks,Wu2018UnsupervisedDiscrimination,Misra2020Self-SupervisedRepresentations}, whereby each training image is treated as its own class. By applying strong data augmentation to these images, and comparing them using a contrastive \cite{Gutmann2010Noise-contrastiveModels,Tian2019ContrastiveCoding,Henaff2019Data-EfficientCoding} loss, a model can learn features which are resilient to various changes in view. The main difficulty in instance discrimination lies in approximating the loss over all instances, as it becomes intractable for large datasets. This leads to metric learning methods which require large numbers of pairwise comparisons. The scaling problem that still remains has been tackled by using memory banks of features \cite{Wu2018UnsupervisedDiscrimination}, momentum encoders \cite{He2019MomentumLearning} or very large batches \cite{Chen2020ARepresentations}. On the other side, clustering-based approaches \cite{Caron2018DeepFeatures,Asano2020Self-labellingLearning} compare groups of images with similar features, sidestepping the intractability of instance discrimination. The problem here instead is computing the cluster assignments over the entire training set. These approaches therefore tend to focus on ways of performing this assignment online \cite{Zhan2020OnlineLearning,Caron2020UnsupervisedAssignments}. Among recent methods, BYOL stands out as one which does not directly use either a contrastive or clustering approach, but as noted by \cite{Tian2020UnderstandingNetworks}, an implicit contrastive loss term is created by their use of batch normalisation. In this paper, we evaluate methods using all of the above approaches, investigating the effect of training objective on transfer performance and representation quality.

\begin{table*}[t]
    \centering
    \caption{Few-shot transfer (5-way 20-shot) of pre-trained models using prototypical networks on popular recognition datasets. Results style: \textbf{best}, \underline{second best}.}
    \label{tab:few_shot_small_domain_shift}
    \resizebox{\textwidth}{!}{%

    \begin{tabular}{lcccccccccc}
    \toprule
    {} &           Aircraft &         Caltech101 &               Cars &            CIFAR10 &           CIFAR100 &                DTD &            Flowers &               Food &               Pets &             SUN397 \\
    \midrule
    InsDis         &              48.67 $\pm$ 0.93 &              89.20 $\pm$ 0.50 &              55.18 $\pm$ 0.69 &              70.16 $\pm$ 0.56 &              75.17 $\pm$ 0.68 &              82.02 $\pm$ 0.50 &              93.76 $\pm$ 0.36 &              70.67 $\pm$ 0.64 &              82.96 $\pm$ 0.57 &              90.81 $\pm$ 0.43 \\
    MoCo-v1        &              48.76 $\pm$ 0.93 &              91.45 $\pm$ 0.43 &              53.04 $\pm$ 0.70 &              66.74 $\pm$ 0.55 &              72.68 $\pm$ 0.70 &              83.08 $\pm$ 0.50 &              93.60 $\pm$ 0.35 &              71.21 $\pm$ 0.65 &              83.68 $\pm$ 0.58 &              90.89 $\pm$ 0.45 \\
    PCL-v1         &              43.31 $\pm$ 0.86 &              87.51 $\pm$ 0.49 &              47.44 $\pm$ 0.75 &              68.16 $\pm$ 0.53 &              69.90 $\pm$ 0.75 &              74.41 $\pm$ 0.62 &              82.75 $\pm$ 0.64 &              65.38 $\pm$ 0.69 &              89.90 $\pm$ 0.52 &              86.40 $\pm$ 0.48 \\
    PIRL           &              49.69 $\pm$ 0.92 &              90.41 $\pm$ 0.46 &              55.82 $\pm$ 0.68 &              71.23 $\pm$ 0.55 &              75.99 $\pm$ 0.70 &              81.98 $\pm$ 0.51 &              93.72 $\pm$ 0.35 &              70.09 $\pm$ 0.66 &              83.61 $\pm$ 0.55 &              91.20 $\pm$ 0.45 \\
    PCL-v2         &              37.68 $\pm$ 0.76 &              88.99 $\pm$ 0.45 &              49.46 $\pm$ 0.73 &              78.22 $\pm$ 0.47 &              80.63 $\pm$ 0.59 &              81.22 $\pm$ 0.54 &              91.81 $\pm$ 0.39 &              69.75 $\pm$ 0.66 &              89.17 $\pm$ 0.52 &              89.37 $\pm$ 0.44 \\
    SimCLR-v1      &              53.55 $\pm$ 0.91 &              95.87 $\pm$ 0.28 &              63.95 $\pm$ 0.78 &              78.10 $\pm$ 0.52 &              82.97 $\pm$ 0.59 &              84.24 $\pm$ 0.46 &              95.69 $\pm$ 0.29 &              74.10 $\pm$ 0.61 &              91.90 $\pm$ 0.43 &              93.83 $\pm$ 0.33 \\
    MoCo-v2        &              39.64 $\pm$ 0.77 &              91.87 $\pm$ 0.40 &              57.67 $\pm$ 0.76 &              76.65 $\pm$ 0.48 &              81.30 $\pm$ 0.63 &              84.57 $\pm$ 0.50 &              94.31 $\pm$ 0.33 &              74.39 $\pm$ 0.64 &              91.78 $\pm$ 0.43 &              92.34 $\pm$ 0.39 \\
    SimCLR-v2      &              53.93 $\pm$ 0.94 &              96.97 $\pm$ 0.22 &              64.25 $\pm$ 0.76 &              79.50 $\pm$ 0.53 &              86.33 $\pm$ 0.55 &              86.42 $\pm$ 0.43 &              96.55 $\pm$ 0.24 &              78.88 $\pm$ 0.57 &              92.24 $\pm$ 0.42 &              95.07 $\pm$ 0.30 \\
    SeLa-v2        &              40.75 $\pm$ 0.86 &              92.67 $\pm$ 0.51 &              57.12 $\pm$ 0.77 &              77.67 $\pm$ 0.51 &              82.42 $\pm$ 0.64 &              85.85 $\pm$ 0.45 &              93.86 $\pm$ 0.34 &              77.26 $\pm$ 0.62 &              88.19 $\pm$ 0.51 &              94.50 $\pm$ 0.33 \\
    InfoMin        &              38.64 $\pm$ 0.75 &              89.12 $\pm$ 0.46 &              57.58 $\pm$ 0.79 &              72.90 $\pm$ 0.52 &              77.25 $\pm$ 0.64 &              80.90 $\pm$ 0.53 &              91.60 $\pm$ 0.40 &              73.99 $\pm$ 0.63 &              91.06 $\pm$ 0.45 &              90.39 $\pm$ 0.45 \\
    BYOL           &  \underline{62.65 $\pm$ 0.92} &  \underline{98.38 $\pm$ 0.15} &  \underline{71.01 $\pm$ 0.75} &              78.73 $\pm$ 0.50 &              85.92 $\pm$ 0.56 &     \textbf{87.56 $\pm$ 0.45} &     \textbf{97.88 $\pm$ 0.19} &              80.07 $\pm$ 0.56 &  \underline{95.71 $\pm$ 0.31} &              95.36 $\pm$ 0.29 \\
    DeepCluster-v2 &              54.68 $\pm$ 0.93 &              97.06 $\pm$ 0.22 &              69.50 $\pm$ 0.77 &  \underline{81.08 $\pm$ 0.49} &  \underline{86.52 $\pm$ 0.54} &  \underline{87.56 $\pm$ 0.42} &  \underline{97.51 $\pm$ 0.20} &  \underline{81.69 $\pm$ 0.55} &              93.80 $\pm$ 0.39 &  \underline{96.26 $\pm$ 0.26} \\
    SwAV           &              53.09 $\pm$ 0.89 &              96.82 $\pm$ 0.23 &              67.83 $\pm$ 0.76 &              79.22 $\pm$ 0.50 &              85.24 $\pm$ 0.57 &              87.33 $\pm$ 0.43 &              97.10 $\pm$ 0.23 &              79.07 $\pm$ 0.59 &              93.84 $\pm$ 0.39 &              96.12 $\pm$ 0.27 \\
    \midrule
    Supervised     &     \textbf{68.90 $\pm$ 0.87} &     \textbf{98.51 $\pm$ 0.16} &     \textbf{82.72 $\pm$ 0.65} &     \textbf{84.29 $\pm$ 0.44} &     \textbf{88.89 $\pm$ 0.49} &              86.58 $\pm$ 0.49 &              96.95 $\pm$ 0.25 &     \textbf{82.93 $\pm$ 0.55} &     \textbf{98.25 $\pm$ 0.19} &     \textbf{96.28 $\pm$ 0.27} \\
    \bottomrule
    \end{tabular}

    }
\end{table*}

\keypoint{Prior evaluations and benchmarks} The importance of empirical evaluation of general purpose representation learning is highlighted by the growing number of major evaluation papers in this area \cite{Kornblith2019DoBetter,Goyal2019ScalingLearning,Zhai2019ABenchmark,Kolesnikov2019RevisitingLearning}.
In terms of transfer performance from supervised pre-training, \cite{Kornblith2019DoBetter} proposes a suite of downstream recognition task evaluations and evaluates transfer performance of several supervised models of varying architecture and pre-training details. They find very strong correlations between ImageNet performance and transfer performance on downstream tasks. In contrast, we compare pre-trained models of exactly the same (ResNet-50) architecture, and instead evaluate the impact of the different  training objectives and augmentation strategies used by self-supervised learners; as well as considering a more diverse suite of downstream benchmarks including few-shot recognition, object detection and dense prediction. Our results are more nuanced, with high correlation visible in recognition tasks similar to ImageNet and lower correlation elsewhere. \cite{Goyal2019ScalingLearning} propose a richer range of downstream benchmarks to evaluate self-supervised pre-training, but focus on the impact of different pre-training datasets and CNN architectures. In contrast, we provide the first comprehensive comparison of different self-supervised algorithms, holding architecture and dataset constant.  \cite{Kolesnikov2019RevisitingLearning} compares a few architectures and SSL algorithms on a small number of downstream tasks, and draw observations such as pre-text task performance being uncorrelated with representation performance on ImageNet recognition. In contrast, we evaluate whether performance on the commonly evaluated ImageNet recognition is indicative of in-the-wild performance on diverse downstream datasets and non-recognition tasks. The evaluation in \cite{Zhai2019ABenchmark} finds that self-supervised methods can not beat supervised models. We find that a more recent family of self-supervised learners consistently achieve the highest performances, on recognition, detection, surface normal estimation and semantic segmentation, with the one exception of few-shot recognition on ImageNet-like data.

%% file: sections/3-preliminaries.tex
\section{Preliminaries}

\begin{table}[b]
    \centering
    \caption{Few-shot transfer (5-way 20-shot) of pre-trained models using prototypical networks on CD-FSL. Results style: \textbf{best}, \underline{second best}.}
    \label{tab:few_shot_large_domain_shift}
    \resizebox{\columnwidth}{!}{%

    \begin{tabular}{lcccc}
    \toprule
    {} & CropDiseases & EuroSAT & ISIC & ChestX \\
    \midrule
    InsDis              &              91.95 $\pm$ 0.44 &              86.52 $\pm$ 0.51 &              52.19 $\pm$ 0.53 &              29.13 $\pm$ 0.44 \\
    MoCo-v1             &              92.04 $\pm$ 0.43 &              86.55 $\pm$ 0.51 &     \textbf{53.79 $\pm$ 0.54} &              30.00 $\pm$ 0.43 \\
    PCL-v1              &              80.74 $\pm$ 0.57 &              75.19 $\pm$ 0.67 &              38.01 $\pm$ 0.44 &              25.54 $\pm$ 0.43 \\
    PIRL                &              91.19 $\pm$ 0.49 &              87.06 $\pm$ 0.50 &              53.24 $\pm$ 0.56 &              29.48 $\pm$ 0.45 \\
    PCL-v2              &              92.58 $\pm$ 0.44 &              87.94 $\pm$ 0.40 &              44.40 $\pm$ 0.52 &              28.28 $\pm$ 0.42 \\
    SimCLR-v1           &              94.03 $\pm$ 0.37 &              89.38 $\pm$ 0.40 &              53.00 $\pm$ 0.54 &              30.82 $\pm$ 0.43 \\
    MoCo-v2             &              92.12 $\pm$ 0.46 &              88.92 $\pm$ 0.41 &              52.39 $\pm$ 0.49 &              29.43 $\pm$ 0.45 \\
    SimCLR-v2           &              94.92 $\pm$ 0.34 &              91.05 $\pm$ 0.36 &              53.15 $\pm$ 0.53 &              30.90 $\pm$ 0.44 \\
    SeLa-v2             &              94.75 $\pm$ 0.37 &              88.34 $\pm$ 0.57 &              48.43 $\pm$ 0.54 &              30.43 $\pm$ 0.46 \\
    InfoMin             &              92.34 $\pm$ 0.44 &              86.76 $\pm$ 0.47 &              48.21 $\pm$ 0.54 &              29.48 $\pm$ 0.44 \\
    BYOL                &              96.07 $\pm$ 0.33 &              89.62 $\pm$ 0.39 &  \underline{53.76 $\pm$ 0.55} &              30.71 $\pm$ 0.47 \\
    DeepCluster-v2      &     \textbf{96.63 $\pm$ 0.29} &     \textbf{92.02 $\pm$ 0.37} &              49.91 $\pm$ 0.53 &     \textbf{31.51 $\pm$ 0.45} \\
    SwAV                &  \underline{96.15 $\pm$ 0.31} &  \underline{91.99 $\pm$ 0.36} &              47.08 $\pm$ 0.50 &  \underline{30.91 $\pm$ 0.45} \\
    \midrule
    Supervised          &              93.09 $\pm$ 0.43 &              88.36 $\pm$ 0.43 &              48.79 $\pm$ 0.53 &              29.26 $\pm$ 0.44 \\
    \bottomrule
    \end{tabular}

    }
\end{table}

\keypoint{Representation learning methods}
We consider the following thirteen self-supervised learning methods.
\textbf{Contrastive}: InsDis (also known as NPID) \cite{Wu2018UnsupervisedDiscrimination}, MoCo-v1 \cite{He2019MomentumLearning} and its upgrade MoCo-v2 \cite{Chen2020ImprovedLearning}, PIRL \cite{Misra2020Self-SupervisedRepresentations}, SimCLR-v1 \cite{Chen2020ARepresentations} and SimCLR-v2 \cite{Chen2020BigLearners}, InfoMin \cite{Tian2020WhatLearning} and BYOL \cite{Grill2020BootstrapLearning}.
\textbf{Clustering}: PCL-v1 and PCL-v2 \cite{Li2020PrototypicalRepresentations}, SeLa-v2 \cite{Asano2020Self-labellingLearning,Caron2020UnsupervisedAssignments}, DeepCluster-v2 \cite{Caron2018DeepFeatures,Caron2020UnsupervisedAssignments} and SwAV \cite{Caron2020UnsupervisedAssignments}.

For these methods, we download pre-trained weights of ResNet50($1\times$) \cite{He2016DeepRecognition} models and use the backbone as a feature extractor when transferring to downstream tasks. Additionally, we evaluate a supervised baseline for comparison, a standard pre-trained ResNet50 available from the PyTorch \cite{Paszke2019PyTorch:Library} library. All models have 23.5M parameters in their backbones and were pre-trained on the ImageNet \cite{Deng2009ImageNet:Database} training set, consisting of 1.28M images, and only the supervised baseline used labels. More details of the pre-trained models can be found in Section~\ref{subsec:pretrained_models} of the appendix.

As we cannot control the pre-training setup, there are differences in how long the models were trained for, what data augmentation they applied, what loss they trained with and what additional architectural elements they used. These differences are detailed in Table \ref{tab:training_details} in the appendix. However, all models use the same ResNet50($1\times$) \cite{He2016DeepRecognition} backbone, meaning we can evaluate them in the same way. For a given target dataset we pass the training data through the backbone to obtain feature vectors. On top of the backbone we attach a task-specific head\cut{ which takes the feature as input} to produce label predictions for the target task. When fitting to the target training set we either optimise only the head or finetune the entire network.

%% file: sections/4-experiments.tex
\section{Experiments}
We now thoroughly evaluate our large suite of recent SSL methods on transfer to a variety of downstream domains and tasks. Our evaluation consists of four sets of transfer experiments: (1) many-shot recognition, where a substantial amount of labelled training data is available in the target domain for fitting a classifier, (2) few-shot recognition  where only a few labelled training images are available for each class in the target domain, and two cases of cross-task transfer, (3) object detection and (4) dense prediction, using the two exemplar tasks: surface normal estimation and semantic segmentation. The first two experiments contain some benchmarks with significant amounts of domain-shift compared to the ImageNet source data, while the last two experiments contain task-shift, that may make different demands on the features. For example, detection may require stronger spatial sensitivity of features compared to recognition; and dense prediction may require something closer to spatial equivariance, in contrast to recognition which may benefit from spatial invariance.

\subsection{Many-shot recognition}\label{sec:manyshot}
\keypoint{Experimental setup}
For many-shot recognition, we adopt the benchmark suite proposed in the transfer learning study \cite{Kornblith2019DoBetter}, which includes the target datasets FGVC Aircraft \cite{Maji2013Fine-GrainedAircraft}, Caltech-101 \cite{Fei-Fei2004LearningCategories}, Stanford Cars \cite{Krause2013CollectingCars}, CIFAR-10 \cite{Krizhevsky2009LearningImages}, CIFAR-100 \cite{Krizhevsky2009LearningImages}, DTD \cite{Cimpoi2014DescribingWild}, Oxford 102 Flowers \cite{Nilsback2008AutomatedClasses}, Food-101 \cite{Bossard2014Food-101Forests}, Oxford-IIIT Pets \cite{Parkhi2012CatsDogs}, SUN397 \cite{Xiao2010SUNZoo} and Pascal VOC2007 \cite{Everingham2010TheChallenge}. These datasets cover a wide range of classification tasks, including texture, scene and fine/coarse-grained object classification. While they are all in the `many-shot' regime, they include significant variety in amount of training data (2,000-75,000 images), and cardinality of classification (10-397 classes). We exclude the Birdsnap \cite{Berg2014Birdsnap:Birds} dataset as a significant number of the original images are no longer available at the given URLs. When using these datasets throughout the paper, we will refer to them collectively as the \emph{Kornblith} datasets.

We report results for both linear evaluation and finetuning. For linear, we fit multinomial logistic regression on the extracted features. When finetuning, we train the models for 5,000 steps using SGD with Nesterov momentum. Full details about our fitting, the dataset splits, metrics and preprocessing can be found in Appendix~\ref{subsec:many_shot_details}.

\begin{table}[t]
    \centering
    \caption{Detection transfer from pre-trained models using Faster R-CNN FPN on PASCAL VOC. We train models both with frozen backbones and with all layers finetuned. We report the metrics AP, AP50 and AP75. Results style: \textbf{best}, \underline{second best}.}
    \label{tab:detection_evaluation}
    \resizebox{0.85\linewidth}{!}{%
    
    \begin{tabular}{lccc|ccc}
    \toprule
    {} & \multicolumn{3}{c}{VOC (Frozen)} & \multicolumn{3}{c}{VOC (Finetune)} \\
    {} &                 AP &               AP50 &               AP75 &                 AP &               AP50 &               AP75 \\
    \midrule
    InsDis         &              50.13 &              77.92 &              53.34 &              48.82 &              76.43 &              52.40 \\
    MoCo-v1        &              50.39 &              78.03 &              54.08 &              50.51 &              78.06 &              54.55 \\
    PCL-v1         &              51.05 &              80.16 &              54.36 &  \underline{53.93} &              81.69 &              59.33 \\
    PIRL           &              49.54 &              77.26 &              52.79 &              45.08 &              72.50 &              47.80 \\
    PCL-v2         &              52.45 &              81.22 &              57.13 &              53.92 &  \underline{81.89} &  \underline{59.35} \\
    SimCLR-v1      &              51.94 &              81.19 &              56.49 &              52.19 &              81.36 &              56.92 \\
    MoCo-v2        &  \underline{54.22} &              81.86 &  \underline{59.97} &              44.74 &              72.82 &              47.01 \\
    SimCLR-v2      &     \textbf{54.95} &     \textbf{82.34} &     \textbf{61.18} &              51.42 &              79.40 &              55.89 \\
    SeLa-v2        &              49.66 &              80.63 &              53.15 &              50.41 &              80.55 &              54.35 \\
    InfoMin        &              53.45 &              81.12 &              58.96 &              44.92 &              72.72 &              47.41 \\
    BYOL           &              53.32 &  \underline{82.01} &              58.37 &     \textbf{54.91} &     \textbf{82.57} &     \textbf{60.82} \\
    DeepCluster-v2 &              50.05 &              80.87 &              53.21 &              51.03 &              80.93 &              55.51 \\
    SwAV           &              50.68 &              80.82 &              54.11 &              52.07 &              81.50 &              56.03 \\
    \midrule
    Supervised     &              51.99 &              81.53 &              56.21 &              53.26 &              81.51 &              59.07 \\
    \bottomrule
    \end{tabular}

    }
\end{table}

\keypoint{Results} 
The results can be found in Table \ref{tab:many_shot_evaluation}\footnote{Note that the linear evaluation in \cite{Kornblith2019DoBetter} uses weights from different checkpoints during pre-training, while we only use the final released weights. This explains why our numbers differ on some datasets.}.

\noindent\emph{Linear:} We draw the following observations: (i) On all but one downstream task, the best self-supervised methods outperform supervised pre-training on ImageNet (bottom row). This is notably the case on Aircraft and Cars benchmarks, where the best self-supervised models outperform supervised pre-training by over 10\% absolute performance. Although supervised pre-training is best for within-dataset transfer to ImageNet (leftmost column), this shows that the self-supervised methods are learning a more general purpose feature for diverse downstream tasks. (ii) The recent methods, DeepCluster-v2 \cite{Caron2020UnsupervisedAssignments}, BYOL \cite{Grill2020BootstrapLearning} and SwAV \cite{Caron2020UnsupervisedAssignments} stand out as being regularly highly ranked in each case.

\noindent\emph{Finetuning:} The bottom half of Table \ref{tab:many_shot_evaluation} shows a similar picture. The supervised model is more competitive here, achieving top results on three datasets including Aircraft where its frozen weights under-performed. However, DeepCluster-v2, SwAV and SimCLR-v2 still outperform it overall, confirming that, on the whole, the best self-supervised learners have surpassed supervision for many-shot recognition transfer. We present further discussion about these results in Section~\ref{sec:correl}.

\subsection{Few-shot recognition}\label{sec:fewshot}
\keypoint{Experimental setup}
To evaluate the performance of self-supervised features on downstream tasks in the few-shot regime, we use the same Kornblith datasets as for the many-shot regime, save for the multi-label VOC2007. Additionally, we evaluate on the Broader Study of Cross-Domain Few-Shot Learning (CD-FSL) benchmark introduced by \cite{Guo2020ALearning}. It consists of four datasets that exhibit increasing dissimilarity to natural images, CropDiseases \cite{Mohanty2016UsingDetection}, EuroSAT \cite{Helber2019Eurosat:Classification},
ISIC2018 \cite{Tschandl2018DataLesions,Codella2019SkinISIC} and ChestX \cite{Wang2017ChestX-ray8:Diseases}.

Our evaluation uses a nearest-centroid classifier (also known as Prototypical Networks \cite{Snell2017PrototypicalLearning}) on the features extracted from the ResNet50 backbones. Across the 14 datasets, we consider 5-way 20-shot transfer (with 5-way 5-shot and 5-way 50-shot reported in the appendix). The test set (query set) always has 15 images per class and we perform 600 randomly sampled few-shot episodes and report the average accuracy along with a 95\% confidence interval.

\keypoint{Results}
Table \ref{tab:few_shot_small_domain_shift} shows the results on the Kornblith datasets. We see that: (i) The supervised model dominates in this setting, on all datasets but DTD and Flowers. (ii) It does so by a large margin on Aircraft and Cars (5+\%), in stark contrast to our linear many-shot results above. (iii) The best self-supervised models are BYOL and DeepCluster-v2, followed by SwAV and SimCLR-v2.

The CD-FSL results are shown in Table \ref{tab:few_shot_large_domain_shift}, from which we make the following observations: (i) Across all datasets and evaluation setups several self-supervised models outperform the supervised baseline. (ii) On CropDiseases, the dataset most similar to ImageNet, the standout models are similar to those in the many-shot experiment: DeepCluster-v2, SwAV and BYOL. On EuroSAT, SimCLR-v2 overtakes BYOL in third place after the same top two. (iii) PCL-v1 consistently transfers the worst in the few-shot setting. (iv) On ISIC, the least `object-like' of all the datasets, the ranking of the methods is very different. We present further discussion about these results in Section~\ref{sec:correl}.

Summarising these results, we see that self-supervision still lags behind for low domain shift few-shot transfer while it consistently beats supervision for larger domain shifts.

\subsection{Detection}\label{sec:detection}
\keypoint{Experimental setup}
We evaluate the pre-trained networks on Pascal VOC using Faster R-CNN \cite{Ren2015FasterNetworks} with a Feature Pyramid Network \cite{Lin2017FeatureDetection} backbone. We use the detectron2 \cite{Wu2019Detectron2} framework and base our evaluation on the suggested hyperparameters therein. Training is done on both the trainval07 and the trainval12 datasets and evaluation is done on the test2007 set. We report AP50, the default VOC metric as well as the COCO-style metrics AP and AP75. We evaluate both freezing the backbone (all but the last residual block) and finetuning all layers end-to-end. Full training details can be found in Section \ref{subsec:detection_details} in the appendix.

\keypoint{Results}
The results are presented in Table~\ref{tab:detection_evaluation}, from which we observe that: (i) The best self-supervised models again outperform supervised pre-training as a transfer learning source. (ii) However, the best performing models are now quite different from those in the previous sections (more on this in Section~\ref{sec:correl}) with SimCLR-v2 excelling for a frozen backbone, and BYOL excelling for a finetuned backbone.

Our results are in contrast to the headline claim in \cite{He2018RethinkingPre-training}, which is that ImageNet pre-training is not necessarily useful in transfer to detection tasks. However, this observation in \cite{He2018RethinkingPre-training} was based on the COCO benchmark, and did not hold for their experiments on Pascal VOC. This is most likely due to the lesser number of images and categories in VOC.

\begin{table}[t]
    \centering
    \caption{Surface normal estimation on NYUv2 (left), with mean and median angular error (lower is better) and percentage of pixels within $11.25^\circ$, $22.5^\circ$, and $30^\circ$ degrees of ground truth surface normal (higher is better). Semantic segmentation on ADE20K (right), with the metrics mean intersection over union and pixel accuracy. Results style: \textbf{best}, \underline{second best}.}
    \label{tab:dense}
    \resizebox{\linewidth}{!}{%

    \begin{tabular}{lccccc|cc}
        \toprule
         & \multicolumn{5}{c}{Surface Normal Estimation} & \multicolumn{2}{c}{Semantic Segmentation} \\
         & Mean & Median & $11.25^\circ$ & $22.5^\circ$ & $30^\circ$ & Mean IoU & Accuracy \\
        \midrule
        InsDis              & 32.99             & 27.35             & 23.58             & 43.02             & 53.51 & 0.2742 & 68.03 \\
        MoCo-v1             & 33.69             & 28.63             & 21.51             & 41.07             & 51.87 & 0.2530 & 62.48 \\
        PCL-v1              & 37.90             & 33.58             & 16.73             & 34.96             & 45.43 & \textbf{0.2983} & \textbf{75.00} \\
        PIRL                & 33.16             & 27.66             & 22.24             & 42.41             & 53.12 & 0.2697 & 66.09 \\
        PCL-v2              & 33.98             & 28.67             & 21.95             & 41.21             & 51.76 & 0.2965 & 74.81 \\
        SimCLR-v1           & 30.47             & 23.26             & 28.34             & 48.88             & \underline{59.01} & \underline{0.2966} & 74.83 \\
        MoCo-v2             & 30.49             & 24.19             & 26.59             & 47.43             & 58.03 & 0.2794 & 67.69 \\
        SimCLR-v2           & \textbf{28.77}    & \textbf{21.30}    & \textbf{30.58}    & \textbf{51.87}    & \textbf{62.05} & 0.2960 & \underline{74.90} \\
        SeLa-v2             & 39.57             & 36.10             & 14.56             & 32.49             & 42.51 & 0.2956 & 74.71 \\
        InfoMin             & 32.45             & 26.58             & 23.86             & 44.00             & 54.66 & 0.2944 & 74.78 \\
        BYOL                & 30.56             & \underline{23.12} & \underline{29.23} & \underline{49.10} & \underline{59.01} & 0.2940 & 74.74 \\
        DeepClust.          & \underline{30.19} & 23.54             & 28.44             & 48.42             & 58.76 & 0.2744   & 67.08 \\
        SwAV                & 31.64             & 24.86             & 27.80             & 46.70             & 56.67 & 0.2961 & 74.87 \\
         \midrule
        Supervised          & 33.52             & 27.91             & 24.00             & 42.33             & 52.80 & 0.2563 & 61.83 \\
        \bottomrule
    \end{tabular}

    }
\end{table}

\begin{figure*}[t]
    \centering
    \includegraphics[width=\linewidth]{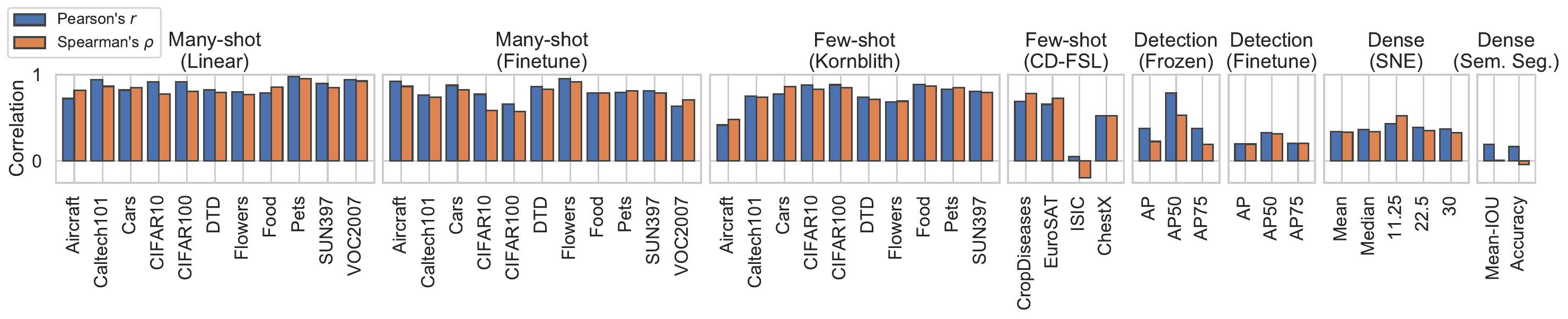}
    \caption{The correlations between ImageNet and downstream transfer performance, showing high correlation for many-shot recognition, but increasingly less so for few-shot, object detection and dense prediction. The blue bars show Pearson's $r$ correlations between logit-transformed ImageNet top-1 accuracy and the transfer performance (which is logit-transformed for metrics bounded between $0$ and $1$, and negated for minimisation metrics). The orange bars show the rank correlation (Spearman's $\rho$).}
    \label{fig:correlations}
\end{figure*}

\subsection{Surface normal estimation}\label{sec:normal_estimation}
\keypoint{Experimental setup} We evaluate the pre-trained features for surface normal estimation on NYUv2~\cite{Silberman2012IndoorImages} (ground-truth from \cite{ladicky2014discriminatively}) as the first exemplar task for dense prediction problems. We train PSPNet models~\cite{zhao2017pyramid} with ResNet50 backbones, as in previous experiments. The performance is measured by the mean and median angular error, as well as the percentage of estimated surface normals within $11.25^\circ$, $22.5^\circ$, and $30^\circ$ of the ground truth.

\keypoint{Results} From the results in Table~\ref{tab:dense}\footnote{Note that our numbers are \emph{not} directly comparable to \cite{Goyal2019ScalingLearning} as they based model (checkpoint) selection on test performance. Given the absence of a validation split for NYUv2, we considered it better practice to train all methods for a fixed number of iterations. As the focus of our benchmark is on comparison across models, this should not be an issue.}, we can see that the best self-supervised models again outperform supervised pre-training for transfer from ImageNet, with SimCLR-v2 winning across the board followed by BYOL. In this case the margins are often substantial with SimCLR-v2 outperforming supervised pre-training by around 4-10\% depending on the metric.

\subsection{Semantic segmentation}\label{sec:sem_seg}
\keypoint{Experimental setup}
The second dense prediction task we consider is semantic segmentation on ADE20K~\cite{zhou2019semantic}. We use the CSAIL Semantic Segmentation framework implementation of UPerNet~\cite{xiao2018unified}, which is based on the Feature Pyramid Network~\cite{Lin2017FeatureDetection} and the Pyramid Pooling Module~\cite{zhao2017pyramid}. We report both the mean intersection over union (IoU) and accuracy.

\keypoint{Results}
We present the results of these experiments in the two rightmost columns of Table~\ref{tab:dense}. The main insights to be gleaned from these performance measurements are: (i) the supervised baseline is among the worst performing methods; (ii) PCL-v1 achieves the top results, while it consistently performed poorly in recognition; and (iii) there is only a very slight correlation between the performance of SSL methods on ImageNet recognition and their performance on semantic segmentation.

\subsection{Does better ImageNet performance lead to better performance on downstream tasks?}\label{sec:correl}

As we mentioned in the introduction, a major question we set out to answer is whether ImageNet performance is in general representative of downstream performance on diverse tasks and datasets? This determines whether practitioners can safely select the latest benchmark leading SSL methods for downstream tasks; and influences whether state-of-the-art self-supervised representations are likely to be useful off-the-shelf for practical problems in diverse domains \cite{Raghu2019Transfusion:Imaging,Guo2020ALearning}, or whether practitioners would need to collect domain-specific data for large scale training. It is also indicative of whether pursuing ImageNet recognition performance is the right benchmark for the self-supervision research community, or whether we need a richer set of benchmarks to properly assess the value of self-supervision research progress to the broader vision community.

\keypoint{Analysis} Based on our experiments in Sections~\ref{sec:manyshot}-\ref{sec:sem_seg}, we compute the Pearson and Spearman (rank) correlation coefficients between ImageNet and downstream task performance across all dataset pairs. Detailed performance plots for every dataset are shown in Figs~\ref{fig:transfer_grid}-\ref{fig:transfer_grid_2} in the appendix. From the summary of correlations in Figs~\ref{fig:transfer_full}-\ref{fig:correlations} we can see that: (i) The ImageNet-to-downstream task correlation is generally high for many-shot recognition tasks. (ii) In the case of few-shot recognition, the correlations are fairly strong for low domain shift transfer. For the larger domain shifts in CD-FSL the correlation is weaker, but present for three of the four datasets. It is entirely absent for the ISIC skin lesion benchmark, which is arguably the least ImageNet-like out of the four due to unstructured texture. (Chest Xray dataset is different due to being greyscale, but similar in the presence of structure in the images). (iii) For detection, AP50 is the strongest correlated metric, and frozen fitting correlates stronger than finetuning. (iv) For surface-normal estimation, weak but clear correlation is present across all metrics. (v) For semantic segmentation the correlation is weak and even non-existent for ranks.

Overall we can distill the following take-home messages for practitioners. (1) \emph{For recognition tasks on structured images, one is safe to choose the current benchmark-leading self-supervised representations for direct transfer purposes in either the many-shot or few-shot regime, and this feature may well out-perform supervised transfer from ImageNet with the exception of few-shot on ImageNet-like data.} (2) \emph{For spatially sensitive prediction tasks such as detection and dense prediction, the current SimCLR-v2 and BYOL are good bets and may outperform supervised transfer, but taking the future ImageNet benchmark leader may not necessarily lead to best performance}. (3) \emph{For recognition tasks on unstructured images and textures, there is no clear recipe to choose a self-supervised representation and task-specific comparison is required.}

\subsection{Does pre-training strategy influence downstream model calibration?}
As computer vision is deployed in many high-importance real-world applications that are safety critical \cite{Kuper2018TowardNetworks}, \cut{or influence individuals well-being \cite{Su2020DeepReview},} or have potential impact on social fairness \cite{Du2019FairnessPerspective}, the \emph{calibration} \cite{Guo2017OnNetworks} of predictive models is as important as overall accuracy, if not a hard-requirement for system deployment. Mistaken predictions should be flagged as such by low-confidence probabilities, so they can be dealt with by another process. Given the growing social importance of this issue, we also evaluate whether pre-training strategy has an influence on downstream model calibration.

\begin{figure}[b]
    \centering
    \includegraphics[width=\linewidth]{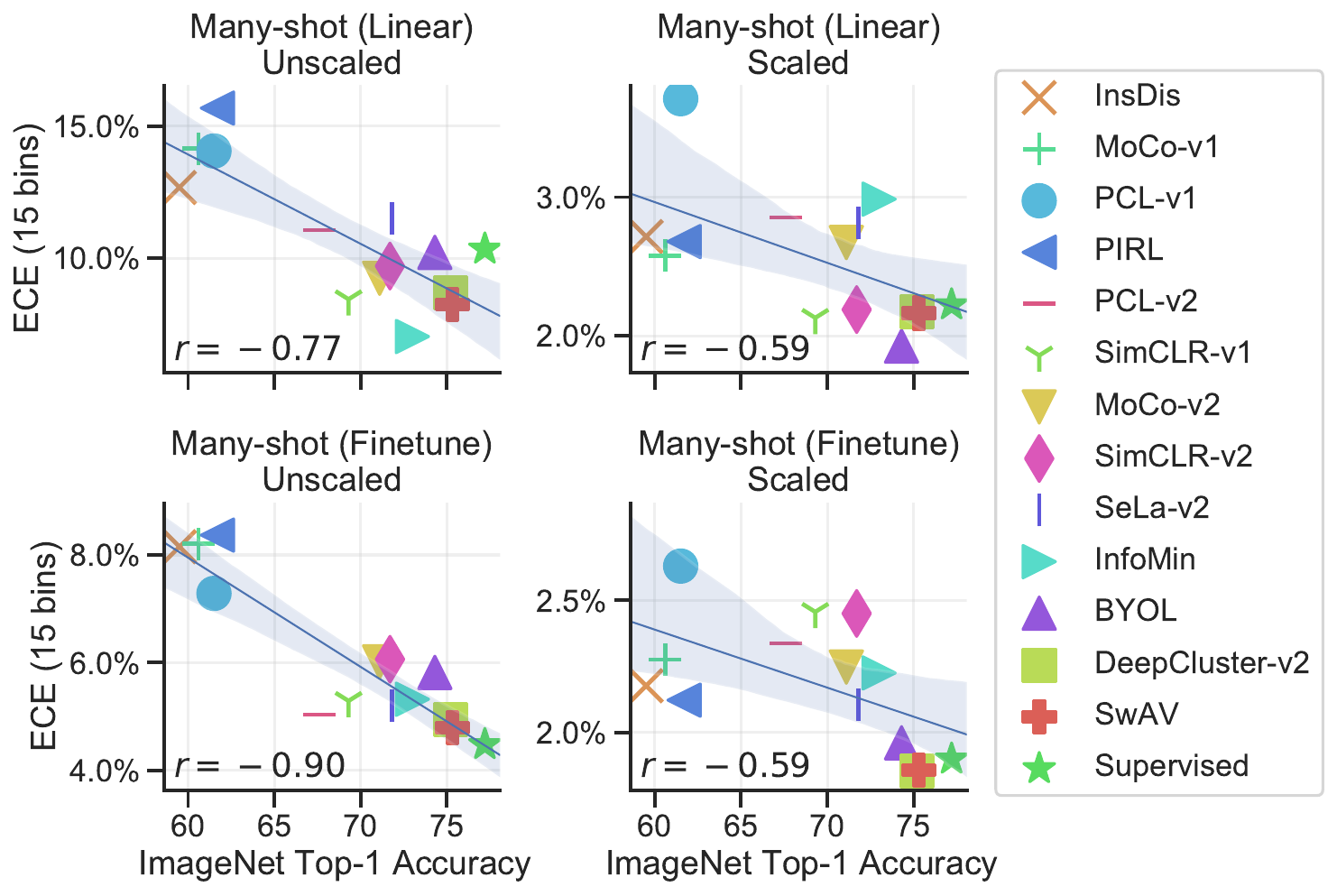}
    \caption{In the linear evaluation setting, many recent self-supervised methods are better calibrated than the supervised baseline. However, after finetuning the supervised model has the best calibration. Overall there is a clear trend that newer SSL models have better calibration (ECE metric, lower is better).}
    \label{fig:calibration}
\end{figure}

We compute the expected calibration error (ECE) \cite{Guo2017OnNetworks} with 15 bins of the models from our two many-shot benchmarks, linear and finetuning. We exclude VOC2007 as it is a multi-label problem. As a simple post-hoc calibration method, we also perform temperature scaling \cite{Guo2017OnNetworks} on the predictions. Figure~\ref{fig:calibration} shows the average ECE for each model over its ImageNet performance both with and without further calibration via temperature scaling.

\keypoint{Analysis}
The overall trend shows better self-supervised methods (as measured on ImageNet accuracy) achieving better calibration. In the unscaled linear case, several SSL models get significantly lower ECE compared to supervision, which also partially holds true after temperature scaling. For unscaled finetuning, the supervised model is the best, though after scaling it is surpassed by DeepCluster-v2 and SwAV. Overall there is a strong inverse correlation of ECE to ImageNet performance -- though less so after temperature scaling -- showing better self-supervised models are better calibrated in downstream transfer.

\subsection{What information is retained in features?}
How to measure what information is retained in CNN features is an open research question in itself \cite{Zeiler2014VisualizingNetworks}. However, to complement our prior performance-driven comparisons, we conduct a preliminary analysis on this topic using the methodology suggested in \cite{Zhao2020WhatLearning}. Specifically, we compare the ability to reconstruct RGB images from the features extracted by our pre-trained models, when using the deep image prior \cite{Ulyanov2018DeepPrior}. This feature inversion algorithm trains an encoder-decoder architecture to produce an image which achieves similar features to the original image when passed through the pre-trained model. We perform image reconstruction from features across all 14 pre-trained models and all 15 unique recognition datasets.

\keypoint{Analysis} To quantify the results we compare: (i) the perceptual difference between original images and reconstructions as measured by \cite{Zhang2018TheMetric}, and (ii) pixel-wise mean squared error between original images and reconstructions. We summarise the results in Figure~\ref{fig:boxplots}, with complete qualitative examples given in Figure~\ref{fig:recon_grid} of the appendix. From the qualitative results we can see that all methods can provide a somewhat recognisable reconstruction, with the noticeable difference that supervised pre-training tends to provide much cleaner colour in the reconstruction. We conjecture that the poor colour fidelity is due to the heavy colour distortions used in the data augmentation of state of the art self-supervised methods leading them to learn colour-invariant features. If so this means that downstream users should be cautious about applying such features to tasks where colour is a critical feature for decision-making. There is a general trend towards stronger methods (in the ImageNet accuracy sense) providing better reconstructions (correlation of -0.69 for perceptual distance computed by the VGG network and for the colour errors, red -0.56, green -0.11, blue -0.22).

\begin{figure}[t]
    \centering
    \includegraphics[width=0.4\columnwidth]{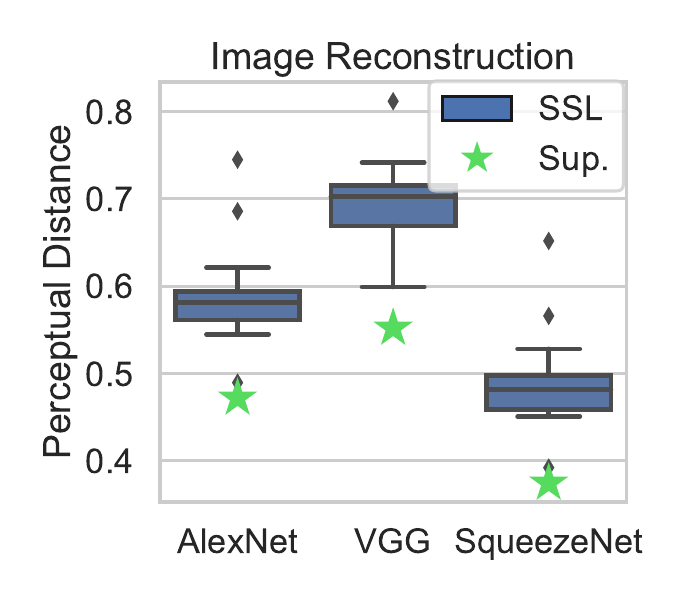} \hspace{-0.3cm}
    \includegraphics[width=0.4\columnwidth]{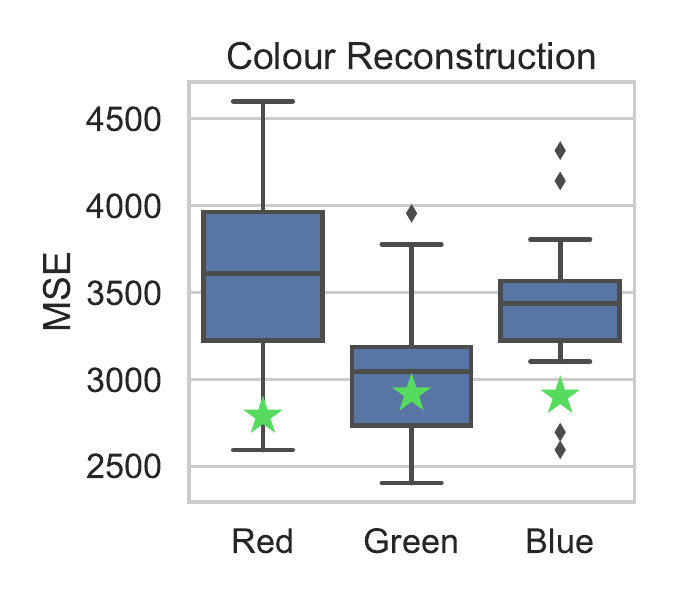} \hspace{-0.3cm}
    \includegraphics[width=0.223\columnwidth]{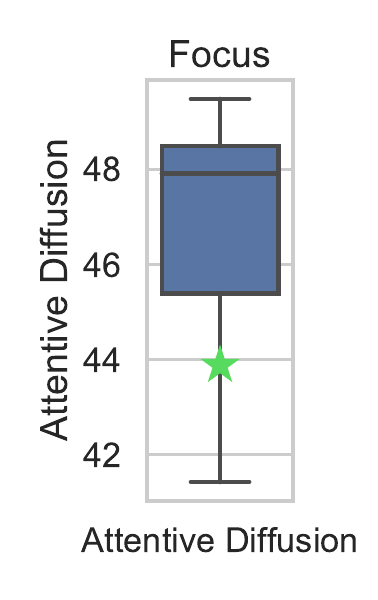}
    \caption{Left: When using features from the supervised baseline (star), the reconstructions are perceptually more similar to the original images compared to the self-supervised models (boxplot). Middle: The supervised model better reconstructs colour information, especially red and blue channels. SSL models likely underperform here because of heavy data augmentation during training. See Fig~\ref{fig:recon_grid} in appendix for reconstructed images. Right: The supervised model has smaller attentive focus compared to SSL models. See Fig~\ref{fig:attention_grid} in appendix for attention maps.}
    \label{fig:boxplots}
\end{figure}

\subsection{Does pre-training strategy influence where downstream networks attend?}
We adapt traditional occlusion-based saliency methods \cite{Zeiler2014VisualizingNetworks} to a task-agnostic setting. By occluding part of the image we compute the distance between the features of the clean and occluded images. As we pass the occlusion mask over the image we compute the average feature distance for each pixel. The larger the value for a given pixel, the more the feature changes if that pixel is occluded in the input, indicating the network is highly sensitive to this region. More details can be found in Section~\ref{subsec:saliency_details} of the appendix.

\keypoint{Analysis} We summarise the results quantitatively in Figure~\ref{fig:boxplots}, with complete qualitative examples given in Figure~\ref{fig:attention_grid} of the appendix. From the qualitative results, some notable observations are that on the aircraft image, the supervised baseline attends to mainly the sky, while the self-supervised ones focus on the actual aircraft. This explains why the supervised model performed so poorly at this fine-grained classification task earlier, as it fails to focus on the details of the aircraft. Overall, there is a trend that the supervised model attends to smaller regions than the self-supervised models. This is summarised quantitatively in Figure~\ref{fig:boxplots}, which reports attentive diffusion/focus in terms of the percentage of the attention map with values above its mean. The correlation with ImageNet performance here is very low at 0.09, but the correlation with average transfer performance (many-shot linear) is significantly higher at 0.38, suggesting that a larger attentive region helps in transfer to recognition tasks. Overall we consider these results to be reflective of widely reported \cite{Zagoruyko2017PayingTransfer} attentive overfitting of supervised learning models, which self-supervised learners seem less vulnerable to, and which may contribute to their superior performance in most recognition tasks and superior calibration for un-tuned backbones.

%% file: sections/5-discussion.tex
\section{Discussion}
We have conducted the first thorough and up-to-date empirical evaluation of state of the art SSL performance when applied to diverse downstream tasks, a comparison that has been missing in the literature until now. Our evaluation showed that: (1) The best self-supervised methods today can usually outperform supervised pre-training as a source of knowledge transfer, an exciting milestone for the field that has long been speculated on, but now clearly confirmed. (2) Performance of self-supervised representations on ImageNet is reassuringly broadly representative of downstream performance on natural image recognition tasks, confirming the relevance of this metric for research. (3) However, ImageNet performance is not reliably representative of downstream performance on unstructured image recognition, or other spatially sensitive tasks such as detection, surface normal prediction and semantic segmentation. Thus the vision of a `universal' pre-trained feature with best performance on diverse downstream tasks is yet to be realised. Furthermore, SSL researchers should adopt a wider range of benchmarks to better impact the broader computer vision community.

There are several limitations of our current study. Most notably, we were not able to compare the value of self-supervised representations transferred from ImageNet to domain-specific self-supervised representations trained on each target dataset. This would answer the important question of whether domain-specific SSL is worthwhile, and if ImageNet can provide truly generic features. This is an important but complex question to answer given the different training protocols of existing methods and diversity of downstream datasets, so we leave this to future work.

%% file: sections/7-appendix.tex
\appendix

\begin{table*}[t]
    \centering
    \caption{5-way 5-shot transfer on the Kornblith datasets. We report the average accuracy and 95\% confidence interval over 600 test episodes. Results style: \textbf{best}, \underline{second best}.}
    \label{tab:five_shot_kornblith}
    \resizebox{0.95\linewidth}{!}{%
    
    \begin{tabular}{lcccccccccc}
    \toprule
    {} &                      Aircraft &                    Caltech101 &                          Cars &                       CIFAR10 &                      CIFAR100 &                           DTD &                       Flowers &                          Food &                          Pets &                        SUN397 \\
    \midrule
    InsDis         &              42.59 $\pm$ 0.90 &              83.31 $\pm$ 0.65 &              46.42 $\pm$ 0.72 &              62.64 $\pm$ 0.64 &              68.06 $\pm$ 0.76 &              73.74 $\pm$ 0.67 &              89.55 $\pm$ 0.53 &              61.50 $\pm$ 0.75 &              73.21 $\pm$ 0.68 &              84.77 $\pm$ 0.60 \\
    MoCo-v1        &              42.74 $\pm$ 0.94 &              86.98 $\pm$ 0.57 &              44.63 $\pm$ 0.69 &              60.07 $\pm$ 0.64 &              66.10 $\pm$ 0.79 &              74.98 $\pm$ 0.70 &              89.13 $\pm$ 0.53 &              62.45 $\pm$ 0.78 &              74.68 $\pm$ 0.69 &              85.14 $\pm$ 0.57 \\
    PCL-v1         &              39.49 $\pm$ 0.87 &              84.35 $\pm$ 0.60 &              40.59 $\pm$ 0.76 &              62.75 $\pm$ 0.63 &              64.09 $\pm$ 0.79 &              64.48 $\pm$ 0.78 &              77.25 $\pm$ 0.75 &              57.45 $\pm$ 0.83 &              85.51 $\pm$ 0.64 &              80.89 $\pm$ 0.62 \\
    PIRL           &              42.91 $\pm$ 0.93 &              85.04 $\pm$ 0.62 &              46.87 $\pm$ 0.74 &              64.39 $\pm$ 0.63 &              69.32 $\pm$ 0.76 &              72.80 $\pm$ 0.69 &              89.52 $\pm$ 0.51 &              61.32 $\pm$ 0.77 &              74.05 $\pm$ 0.69 &              85.03 $\pm$ 0.59 \\
    PCL-v2         &              34.36 $\pm$ 0.75 &              86.33 $\pm$ 0.54 &              42.57 $\pm$ 0.70 &              70.96 $\pm$ 0.59 &              74.10 $\pm$ 0.69 &              72.84 $\pm$ 0.74 &              87.52 $\pm$ 0.52 &              61.00 $\pm$ 0.78 &              85.16 $\pm$ 0.66 &              84.80 $\pm$ 0.57 \\
    SimCLR-v1      &              48.11 $\pm$ 0.98 &              94.10 $\pm$ 0.36 &              53.46 $\pm$ 0.80 &              70.65 $\pm$ 0.66 &              77.10 $\pm$ 0.70 &              76.71 $\pm$ 0.65 &              93.10 $\pm$ 0.38 &              65.13 $\pm$ 0.77 &              86.52 $\pm$ 0.58 &              89.71 $\pm$ 0.47 \\
    MoCo-v2        &              35.97 $\pm$ 0.80 &              90.14 $\pm$ 0.48 &              49.55 $\pm$ 0.80 &              69.47 $\pm$ 0.62 &              75.62 $\pm$ 0.70 &              78.08 $\pm$ 0.67 &              91.12 $\pm$ 0.46 &              66.34 $\pm$ 0.80 &              87.91 $\pm$ 0.59 &              89.18 $\pm$ 0.48 \\
    SimCLR-v2      &              47.12 $\pm$ 0.96 &              94.92 $\pm$ 0.34 &              52.64 $\pm$ 0.77 &              71.90 $\pm$ 0.61 &              79.71 $\pm$ 0.66 &              79.06 $\pm$ 0.63 &              93.83 $\pm$ 0.37 &              69.85 $\pm$ 0.74 &              86.29 $\pm$ 0.58 &              90.99 $\pm$ 0.45 \\
    SeLa-v2        &              36.35 $\pm$ 0.77 &              89.85 $\pm$ 0.53 &              47.99 $\pm$ 0.78 &              71.27 $\pm$ 0.59 &              76.29 $\pm$ 0.72 &              77.81 $\pm$ 0.62 &              90.11 $\pm$ 0.51 &              67.69 $\pm$ 0.77 &              81.36 $\pm$ 0.67 &              90.80 $\pm$ 0.46 \\
    InfoMin        &              35.06 $\pm$ 0.75 &              87.03 $\pm$ 0.53 &              49.67 $\pm$ 0.79 &              67.28 $\pm$ 0.62 &              71.72 $\pm$ 0.72 &              73.43 $\pm$ 0.75 &              87.53 $\pm$ 0.57 &              65.95 $\pm$ 0.77 &              86.98 $\pm$ 0.57 &              86.54 $\pm$ 0.55 \\
    BYOL           &  \underline{53.88 $\pm$ 0.99} &  \underline{96.84 $\pm$ 0.28} &  \underline{58.77 $\pm$ 0.81} &              70.59 $\pm$ 0.62 &              79.19 $\pm$ 0.68 &     \textbf{81.33 $\pm$ 0.59} &     \textbf{96.06 $\pm$ 0.30} &              71.39 $\pm$ 0.72 &  \underline{92.20 $\pm$ 0.46} &              91.63 $\pm$ 0.43 \\
    DeepCluster-v2 &              47.73 $\pm$ 0.97 &              94.75 $\pm$ 0.35 &              58.17 $\pm$ 0.82 &  \underline{74.47 $\pm$ 0.61} &  \underline{80.52 $\pm$ 0.65} &              78.79 $\pm$ 0.59 &  \underline{95.44 $\pm$ 0.32} &  \underline{72.71 $\pm$ 0.72} &              89.13 $\pm$ 0.56 &              92.95 $\pm$ 0.41 \\
    SwAV           &              46.22 $\pm$ 0.91 &              94.43 $\pm$ 0.37 &              56.08 $\pm$ 0.82 &              72.73 $\pm$ 0.62 &              79.32 $\pm$ 0.67 &              79.80 $\pm$ 0.57 &              94.55 $\pm$ 0.37 &              69.65 $\pm$ 0.73 &              88.76 $\pm$ 0.56 &  \underline{93.00 $\pm$ 0.42} \\
    \midrule
    Supervised     &     \textbf{58.35 $\pm$ 0.96} &     \textbf{97.61 $\pm$ 0.24} &     \textbf{73.68 $\pm$ 0.84} &     \textbf{77.50 $\pm$ 0.55} &     \textbf{83.74 $\pm$ 0.61} &  \underline{80.83 $\pm$ 0.59} &              94.19 $\pm$ 0.41 &     \textbf{76.23 $\pm$ 0.71} &     \textbf{97.45 $\pm$ 0.28} &     \textbf{93.78 $\pm$ 0.38} \\
    \bottomrule
    \end{tabular}

    }
\end{table*}

\begin{table*}[t]
    \centering
    \caption{5-way 50-shot transfer on the Kornblith datasets, apart from Caltech101, Cars and Flowers which do not have enough images per class for this setup. We report the average accuracy and 95\% confidence interval over 600 test episodes. Results style: \textbf{best}, \underline{second best}.}
    \label{tab:fifty_shot_kornblith}
    \resizebox{0.68\linewidth}{!}{%
    
    \begin{tabular}{lccccccc}
    \toprule
    {} &                      Aircraft &                       CIFAR10 &                      CIFAR100 &                           DTD &                          Food &                          Pets &                        SUN397 \\
    \midrule
    InsDis         &              51.06 $\pm$ 0.88 &              71.77 $\pm$ 0.52 &              77.57 $\pm$ 0.63 &              83.97 $\pm$ 0.47 &              73.43 $\pm$ 0.63 &              84.78 $\pm$ 0.56 &              92.10 $\pm$ 0.39 \\
    MoCo-v1        &              51.20 $\pm$ 0.89 &              68.22 $\pm$ 0.54 &              75.22 $\pm$ 0.70 &              84.76 $\pm$ 0.49 &              74.19 $\pm$ 0.60 &              85.65 $\pm$ 0.55 &              92.31 $\pm$ 0.38 \\
    PCL-v1         &              44.78 $\pm$ 0.82 &              69.35 $\pm$ 0.53 &              72.07 $\pm$ 0.70 &              77.18 $\pm$ 0.58 &              67.46 $\pm$ 0.67 &              90.76 $\pm$ 0.46 &              87.59 $\pm$ 0.47 \\
    PIRL           &              52.17 $\pm$ 0.88 &              72.23 $\pm$ 0.52 &              78.43 $\pm$ 0.64 &              83.94 $\pm$ 0.51 &              73.05 $\pm$ 0.62 &              85.58 $\pm$ 0.53 &              92.44 $\pm$ 0.39 \\
    PCL-v2         &              38.48 $\pm$ 0.78 &              79.51 $\pm$ 0.45 &              82.86 $\pm$ 0.53 &              83.79 $\pm$ 0.48 &              72.30 $\pm$ 0.65 &              89.96 $\pm$ 0.48 &              90.19 $\pm$ 0.42 \\
    SimCLR-v1      &              55.29 $\pm$ 0.93 &              79.72 $\pm$ 0.49 &              84.43 $\pm$ 0.55 &              86.24 $\pm$ 0.43 &              77.24 $\pm$ 0.59 &              92.83 $\pm$ 0.40 &              94.34 $\pm$ 0.33 \\
    MoCo-v2        &              41.22 $\pm$ 0.79 &              78.01 $\pm$ 0.45 &              83.01 $\pm$ 0.57 &              86.42 $\pm$ 0.46 &              77.17 $\pm$ 0.60 &              92.25 $\pm$ 0.42 &              92.98 $\pm$ 0.36 \\
    SimCLR-v2      &              56.33 $\pm$ 0.91 &              81.36 $\pm$ 0.48 &              87.79 $\pm$ 0.49 &              87.99 $\pm$ 0.42 &              81.65 $\pm$ 0.53 &              93.51 $\pm$ 0.38 &              95.51 $\pm$ 0.28 \\
    SeLa-v2        &              43.04 $\pm$ 0.83 &              79.16 $\pm$ 0.50 &              84.11 $\pm$ 0.59 &              87.77 $\pm$ 0.43 &              80.10 $\pm$ 0.56 &              89.84 $\pm$ 0.44 &              95.11 $\pm$ 0.29 \\
    InfoMin        &              39.91 $\pm$ 0.76 &              74.23 $\pm$ 0.53 &              79.16 $\pm$ 0.57 &              83.09 $\pm$ 0.49 &              76.12 $\pm$ 0.59 &              91.61 $\pm$ 0.42 &              91.05 $\pm$ 0.42 \\
    BYOL           &  \underline{65.69 $\pm$ 0.88} &              80.49 $\pm$ 0.47 &              87.57 $\pm$ 0.50 &              89.12 $\pm$ 0.42 &              83.04 $\pm$ 0.51 &  \underline{96.18 $\pm$ 0.30} &              95.89 $\pm$ 0.26 \\
    DeepCluster-v2 &              57.84 $\pm$ 0.93 &  \underline{82.56 $\pm$ 0.47} &  \underline{88.11 $\pm$ 0.46} &     \textbf{89.34 $\pm$ 0.40} &  \underline{84.38 $\pm$ 0.49} &              94.62 $\pm$ 0.36 &              96.57 $\pm$ 0.24 \\
    SwAV           &              55.88 $\pm$ 0.89 &              80.30 $\pm$ 0.49 &              86.93 $\pm$ 0.51 &  \underline{89.13 $\pm$ 0.41} &              81.94 $\pm$ 0.54 &              94.58 $\pm$ 0.36 &     \textbf{96.64 $\pm$ 0.24} \\
    \midrule
    Supervised     &     \textbf{71.97 $\pm$ 0.83} &     \textbf{85.80 $\pm$ 0.40} &     \textbf{90.24 $\pm$ 0.42} &              88.23 $\pm$ 0.44 &     \textbf{85.26 $\pm$ 0.48} &     \textbf{98.54 $\pm$ 0.16} &  \underline{96.61 $\pm$ 0.24} \\
    \bottomrule
    \end{tabular}

    }
\end{table*}

\begin{table*}[t]
    \centering
    \caption{Few-shot transfer of pre-trained models using prototypical networks. Here, we present few-shot transfer results for 5-way 5-shot and 5-way 50-shot settings on CD-FSL. We report the average accuracy and 95\% confidence interval over 600 test episodes. Results style: \textbf{best}, \underline{second best}.}
    \label{tab:five_and_fifty_shot_cdfsl}
    \resizebox{0.8\linewidth}{!}{%

    \begin{tabular}{lcccccccc}
    \toprule
    {} & \multicolumn{2}{c}{CropDiseases} & \multicolumn{2}{c}{EuroSAT} & \multicolumn{2}{c}{ISIC} & \multicolumn{2}{c}{ChestX} \\
    {} &                        5-shot &                       50-shot &                        5-shot &                       50-shot &                        5-shot &                       50-shot &                        5-shot &                       50-shot \\
    \midrule
    InsDis              &              88.01 $\pm$ 0.58 &              92.70 $\pm$ 0.43 &              81.29 $\pm$ 0.63 &              88.25 $\pm$ 0.47 &              43.90 $\pm$ 0.55 &              55.76 $\pm$ 0.50 &              25.67 $\pm$ 0.42 &              31.77 $\pm$ 0.44 \\
    MoCo-v1             &              87.87 $\pm$ 0.58 &              92.87 $\pm$ 0.42 &              81.32 $\pm$ 0.61 &              87.72 $\pm$ 0.46 &     \textbf{44.42 $\pm$ 0.55} &              56.81 $\pm$ 0.52 &              25.92 $\pm$ 0.45 &              32.74 $\pm$ 0.43 \\
    PCL-v1              &              72.89 $\pm$ 0.69 &              82.83 $\pm$ 0.55 &              66.56 $\pm$ 0.76 &              76.41 $\pm$ 0.63 &              33.21 $\pm$ 0.48 &              39.77 $\pm$ 0.45 &              23.33 $\pm$ 0.40 &              27.40 $\pm$ 0.42 \\
    PIRL                &              86.22 $\pm$ 0.63 &              92.18 $\pm$ 0.44 &              82.14 $\pm$ 0.63 &              88.55 $\pm$ 0.44 &              43.89 $\pm$ 0.54 &  \underline{56.89 $\pm$ 0.52} &              25.60 $\pm$ 0.41 &              31.44 $\pm$ 0.47 \\
    PCL-v2              &              87.57 $\pm$ 0.60 &              93.57 $\pm$ 0.40 &              81.10 $\pm$ 0.54 &              89.23 $\pm$ 0.37 &              37.47 $\pm$ 0.52 &              46.82 $\pm$ 0.46 &              24.87 $\pm$ 0.42 &              30.56 $\pm$ 0.43 \\
    SimCLR-v1           &              90.29 $\pm$ 0.52 &              94.49 $\pm$ 0.37 &              82.78 $\pm$ 0.56 &              90.55 $\pm$ 0.36 &  \underline{43.99 $\pm$ 0.55} &              56.16 $\pm$ 0.53 &              26.36 $\pm$ 0.44 &              33.16 $\pm$ 0.47 \\
    MoCo-v2             &              87.62 $\pm$ 0.60 &              93.61 $\pm$ 0.40 &              84.15 $\pm$ 0.52 &              89.83 $\pm$ 0.37 &              42.60 $\pm$ 0.55 &              55.68 $\pm$ 0.53 &              25.26 $\pm$ 0.44 &              32.20 $\pm$ 0.43 \\
    SimCLR-v2           &              90.80 $\pm$ 0.52 &              95.80 $\pm$ 0.29 &              86.45 $\pm$ 0.49 &              92.07 $\pm$ 0.30 &              43.66 $\pm$ 0.58 &              56.83 $\pm$ 0.54 &              26.34 $\pm$ 0.44 &              33.23 $\pm$ 0.47 \\
    SeLa-v2             &              90.96 $\pm$ 0.54 &              95.40 $\pm$ 0.33 &              84.56 $\pm$ 0.57 &              88.51 $\pm$ 0.59 &              39.97 $\pm$ 0.55 &              51.31 $\pm$ 0.52 &              25.60 $\pm$ 0.44 &              32.81 $\pm$ 0.44 \\
    InfoMin             &              87.77 $\pm$ 0.61 &              92.93 $\pm$ 0.40 &              81.68 $\pm$ 0.59 &              87.61 $\pm$ 0.43 &              39.03 $\pm$ 0.55 &              51.58 $\pm$ 0.51 &              25.78 $\pm$ 0.44 &              31.58 $\pm$ 0.44 \\
    BYOL                &              92.71 $\pm$ 0.47 &              96.69 $\pm$ 0.27 &              83.64 $\pm$ 0.54 &              90.46 $\pm$ 0.35 &              43.09 $\pm$ 0.56 &     \textbf{58.03 $\pm$ 0.52} &              26.39 $\pm$ 0.43 &     \textbf{34.17 $\pm$ 0.45} \\
    DeepCluster-v2      &     \textbf{93.63 $\pm$ 0.44} &     \textbf{97.04 $\pm$ 0.27} &     \textbf{88.39 $\pm$ 0.49} &  \underline{93.07 $\pm$ 0.31} &              40.73 $\pm$ 0.59 &              53.65 $\pm$ 0.54 &  \underline{26.51 $\pm$ 0.45} &     \textbf{34.17 $\pm$ 0.48} \\
    SwAV                &  \underline{93.49 $\pm$ 0.46} &  \underline{96.72 $\pm$ 0.28} &  \underline{87.29 $\pm$ 0.54} &     \textbf{93.36 $\pm$ 0.31} &              39.66 $\pm$ 0.54 &              51.10 $\pm$ 0.50 &     \textbf{26.54 $\pm$ 0.48} &  \underline{33.86} $\pm$ 0.46 \\
    \midrule
    Supervised          &              89.37 $\pm$ 0.55 &              94.32 $\pm$ 0.36 &              83.81 $\pm$ 0.55 &              89.62 $\pm$ 0.37 &              39.38 $\pm$ 0.58 &              52.54 $\pm$ 0.56 &              25.22 $\pm$ 0.41 &              32.34 $\pm$ 0.45 \\
    \bottomrule
    \end{tabular}

    }
\end{table*}

\begin{table*}[t]
    \centering
    \caption{Numerical values for the results presented in Figs~\ref{fig:calibration}-\ref{fig:boxplots}. Columns 1-4: Expected calibration error (ECE) using 15 bins for unscaled models and models further calibrated using temperature scaling. Columns 5-7: Average perceptual distance computed on reconstructed images, using three different measures of the Learned Perceptual Image Patch Similarity (LPIPS) metric \cite{Zhang2018TheMetric}. Columns 8-10: Mean squared errors between the colour channels of reconstructed and original images. Column 11: Attentive diffusion measured as the percentage of attention values above the mean attention over an image. Higher value means wider attention. Results style: \textbf{lowest}, \underline{second lowest}.}
    \label{tab:qualitative_results}

    \resizebox{0.8\linewidth}{!}{%
    \begin{tabular}{lrrrr|rrr|rrr|r}
    \toprule
    {} & \multicolumn{2}{c}{Many-shot (Linear)} & \multicolumn{2}{c}{Many-shot (Finetune)} &  \multicolumn{3}{c}{Perceptual Distance} & \multicolumn{3}{c}{Colour Error} &  Attention \\
    {} &  \multicolumn{1}{c}{Unscaled} &  \multicolumn{1}{c}{Scaled} &  \multicolumn{1}{c}{Unscaled} &  \multicolumn{1}{c}{Scaled} &  \multicolumn{1}{c}{AlexNet} &  \multicolumn{1}{c}{VGG} &  \multicolumn{1}{c}{SqueezeNet} &     \multicolumn{1}{c}{Red} &   \multicolumn{1}{c}{Green} &    \multicolumn{1}{c}{Blue} &  \multicolumn{1}{c}{Diffusion} \\
    \midrule
    InsDis      &               12.68 &                             2.72 &                  8.15 &                               2.18 &     0.58 & 0.71 &        0.48 & 3971 & 2734 & 3394 &                48.48 \\
    MoCo-v1     &               14.15 &                             2.58 &                  8.21 &                               2.28 &     0.62 & 0.74 &        0.53 & 4073 & 3044 & 3512 &                47.92 \\
    PCL-v1      &               14.06 &                             3.71 &                  7.29 &                               2.63 &     0.74 & 0.81 &        0.65 & 4598 & 3954 & 4141 &                \textbf{41.43} \\
    PIRL        &               15.68 &                             2.68 &                  8.37 &                               2.12 &     0.59 & 0.72 &        0.50 & 3607 & 3070 & 3435 &                48.12 \\
    PCL-v2      &               11.07 &                             2.85 &                  5.04 &                               2.34 &     0.56 & 0.66 &        0.47 & 3008 & 2807 & 3101 &                43.91 \\
    SimCLR-v1   &                8.45 &                             \underline{2.13} &                  5.29 &                               2.46 &     0.56 & 0.70 &        0.47 & 3224 & 2667 & 3223 &                46.07 \\
    MoCo-v2     &                9.25 &                             2.67 &                  6.01 &                               2.25 &     0.54 & 0.67 &        0.45 & 3179 & \underline{2514} & \underline{2695} &                45.39 \\
    SimCLR-v2   &                9.71 &                             2.19 &                  6.06 &                               2.45 &     0.55 & 0.68 &        0.46 & 3655 & 2855 & 3404 &                47.91 \\
    SeLa-v2     &               11.52 &                             2.81 &                  5.20 &                               2.10 &     0.69 & 0.72 &        0.57 & 3962 & 3775 & 4315 &                47.68 \\
    InfoMin     &                \textbf{7.05} &                             2.99 &                  5.32 &                               2.23 &     \underline{0.49} & \underline{0.60} &        \underline{0.39} & \textbf{2592} & \textbf{2403} & \textbf{2594} &                \underline{43.73} \\
    BYOL        &               10.23 &                             \textbf{1.93} &                  5.82 &                               1.96 &     0.59 & 0.71 &        0.48 & 3765 & 3268 & 3471 &                48.81 \\
    DeepCluster-v2 &                8.69 &                             2.17 &                  4.94 &                               \textbf{1.85} &     0.58 & 0.67 &        0.48 & 3527 & 3170 & 3804 &                48.69 \\
    SwAV        &                \underline{8.25} &                             2.16 &                  \underline{4.80} &                               \underline{1.86} &     0.57 & 0.67 &        0.46 & 3560 & 3186 & 3565 &                49.47 \\
    \midrule
    Supervised  &               10.35 &                             2.22 &                  \textbf{4.48} &                               1.90 &     \textbf{0.47} & \textbf{0.55} &        \textbf{0.37} & \underline{2788} & 2917 & 2903 &                43.88 \\
    \midrule \midrule
    Correlation to ImageNet & -0.77 & -0.59 & -0.90 & -0.59 & -0.51 & -0.69 & -0.57 & -0.56 & -0.11 & -0.22 & 0.09 \\
    \bottomrule
    \end{tabular}

    }
\end{table*}

\begin{table*}[t]
\centering
\caption{Training details as reported by original authors for all models used in this paper. Asterisks (*) note models we obtain from PyContrast instead of original authors.}
\label{tab:training_details}
\resizebox{\linewidth}{!}{%
\begin{tabular}{@{}lcccccccccccccc@{}}
\toprule
\textbf{}           & Epochs       & Batch size    & Target net & Mom. enc.  & Mem. bank  & Proj. head & Jigsaw     & Grayscale  & Colour jitter & Solarize   & Blur       & Random crop & Horiz. flip & Normalize \\
\midrule
InsDis*             & 200          & 256           &            &            & \checkmark &            &            & \checkmark & \checkmark    &            &            & \checkmark  & \checkmark      & \checkmark                    \\
MoCo-v1             & 200          & 256           &            & \checkmark &            &            &            & \checkmark & \checkmark    &            &            & \checkmark  & \checkmark      & \checkmark                    \\
PCL-v1              & 200          & 256           &            & \checkmark &            &            &            & \checkmark & \checkmark    &            &            & \checkmark  & \checkmark      & \checkmark                    \\
PIRL*               & 200          & 1024          &            &            & \checkmark &            & \checkmark &            & \checkmark    &            &            & \checkmark  & \checkmark      & \checkmark                    \\
PCL-v2              & 200          & 256           &            & \checkmark &            & \checkmark &            & \checkmark & \checkmark    &            & \checkmark & \checkmark  & \checkmark      & \checkmark                    \\
SimCLR-v1           & 1000         & 4096          &            &            &            & \checkmark &            & \checkmark & \checkmark    &            & \checkmark & \checkmark  & \checkmark      & \multicolumn{1}{l}{}          \\
MoCo-v2             & 800          & 256           &            & \checkmark &            & \checkmark &            & \checkmark & \checkmark    &            & \checkmark & \checkmark  & \checkmark      & \checkmark                    \\
SimCLR-v2           & 800          & 4096          &            & \checkmark &            & \checkmark &            & \checkmark & \checkmark    &            & \checkmark & \checkmark  & \checkmark      & \multicolumn{1}{l}{}          \\
SeLa-v2             & 400          & 4096          &            &            & \checkmark & \checkmark &            & \checkmark & \checkmark    &            & \checkmark & multi       & \checkmark      & \checkmark                    \\
InfoMin             & 800          & 256           &            & \checkmark &            & \checkmark & \checkmark & \checkmark & \checkmark    &            & \checkmark & \checkmark  & \checkmark      & \checkmark                    \\
BYOL                & 1000         & 4096          & \checkmark &            &            & \checkmark &            & \checkmark & \checkmark    & \checkmark & \checkmark & \checkmark  & \checkmark      & \checkmark                    \\
DeepCluster-v2      & 800 & 4096 &            &            & \checkmark & \checkmark &            & \checkmark & \checkmark    &            & \checkmark & multi       & \checkmark      & \checkmark                    \\
SwAV                & 800          & 4096          &            &            &            & \checkmark &            & \checkmark & \checkmark    &            & \checkmark & multi       & \checkmark      & \checkmark                    \\ \midrule
Supervised          & 120          & 256           &            &            &            &            &            &            & PCA           &            &            & \checkmark  & \checkmark      & \checkmark                    \\
\bottomrule
\end{tabular}%
}
\end{table*}

\section{Appendix} \label{appendix}

\subsection{Pre-trained models} \label{subsec:pretrained_models}
Note that for InsDis we use the model weights provided by the PyContrast GitHub repository which report higher ImageNet top-1 accuracy than originally reported (59.5 vs 54.0). As weights are not available for PIRL we likewise, take the ones provided by PyContrast which reports a slightly lower ImageNet accuracy of 61.7 (compared to 63.6). All other models are obtained from the original authors. We use the PyTorch framework in our code and therefore convert some of the models from their TensorFlow checkpoints. For most models we normalise the inputs by the mean and standard deviation on the ILSVRC12 train set, apart from SimCLR-v1/v2 which do not expect normalised inputs.

\subsection{Many-shot evaluation details} \label{subsec:many_shot_details}
The top-1 accuracy metric is reported on Food-101, CIFAR-10, CIFAR-100, SUN397, Stanford Cars, and DTD, mean per-class accuracy on FGVC Aircraft, Oxford-IIIT Pets, Caltech-101, and Oxford 102 Flowers and the 11-point mAP metric from \cite{Everingham2010TheChallenge} on Pascal VOC 2007. On Caltech-101 we randomly select 30 images per class to form the training set and we test on the rest. We use the first train/test split defined in DTD and SUN397. On FGVC Aircraft, Pascal VOC2007, DTD, and Oxford 102 Flowers we use the validation sets defined by the authors, and on the other datasets we randomly select 20\% of the training set to form the validation set. The optimal hyperparameters were selected on the validation set, after which we retrained the model on all training and validation images. Finally, the accuracy is computed on the test set.

\keypoint{Linear}
We fit a multinomial logistic regression model on the extracted features of dimensionality 2048 from the frozen backbones. No augmentation was used and the images were resized to 224 pixels along the shorter side using bicubic resampling, followed by a center crop of 224 $\times$ 224. We select the $\ell 2$ regularisation constant on the validation set over 45 logarithmically spaced values between $10^{-6}$ and $10^5$. The model is optimised using L-BFGS \cite{Liu1989OnOptimization} on the softmax cross-entropy objective. As Pascal VOC2007 is a multi-label task, we fit one binary classifier for each class.

\keypoint{Finetuning}
We finetune the models following the protocol of \cite{Chen2020ARepresentations} with minor modifications. We train for 5000 steps with a batch size of 64. The optimiser is SGD with Nesterov momentum and a momentum parameter of 0.9. \cut{The momentum parameter for the batch normalization layers is set to $max(1 - \frac{10}{s}, 0.9)$ where $s$ is the number of steps per epoch. }The learning rate follows a cosine annealing schedule without restarts, and the initial learning rate is chosen from a grid of 4 logarithmically spaced values between 0.0001 and 0.1. The weight decay is similarly chosen from a grid of 4 logarithmically spaced values between $10^{-6}$ and $10^{-3}$, along with no weight decay. These weight decay values are divided by the learning rate. We select the data augmentation from: random crop with resize and flip, or simply a center crop.

\subsection{Few-shot evaluation details} \label{subsec:few_shot_details}
For each few-shot learning episode we sample images from the combined sets of train, validation and test images. We fit a nearest centroid classifier on the extracted features of dimensionality 2048 from the frozen backbones. No augmentation was used and the images were resized to 224 pixels along the shorter side using bicubic resampling, followed by a center crop of 224 $\times$ 224. The fitted model is evaluated using 15 query images in each episode and the reported accuracies and errors are computed from 600 total episodes. In addition to the 20-shot results presented in the paper, we also report 5-shot and 50-shot results in Tables~\ref{tab:five_shot_kornblith}, \ref{tab:fifty_shot_kornblith} and \ref{tab:five_and_fifty_shot_cdfsl}. Note that in the original CD-FSL benchmark \cite{Guo2020ALearning}, models are only allowed to pre-train on mini-ImageNet and not the full version, so our results are not comparable to those of the original authors.

\subsection{Detection evaluation details} \label{subsec:detection_details}
We train the detectors on the VOC 2007 and 2012 trainval sets, and test on VOC 2007 test. When evaluating frozen backbones, we freeze all but the final residual block of the ResNets. In the full finetuning setup, we let the entire network be trainable. We extract features from the backbone using a Feature Pyramid Network \cite{Lin2017FeatureDetection} architecture and attach a Faster R-CNN \cite{Ren2015FasterNetworks} detector head to produce predictions. During training, the images are resized so the shorter side is one of [480, 512, 544, 576, 608, 640, 672, 704, 736, 768, 800] and during testing to 800 pixels. The models are trained for 144k iterations with a 100 iteration warm-up to an initial learning rate of 0.0025 which is decayed by a factor of 10 at iterations 96k and 128k. The batch size is 2 and we used a single GPU per model. Any other details of training uses the default values of the detectron2 \cite{Wu2019Detectron2} framework.

\subsection{Surface normal estimation evaluation details} \label{subsec:sne_details}
We use the implementation of~\cite{Goyal2019ScalingLearning}, which is based on~\cite{zhou2019semantic}. Each model is trained for 150 epochs, with the full backbone frozen. We use stochastic gradient descent with a momentum of $0.9$, batch size of 4 and set the learning rate according to $(1 - \frac{t}{T})^{0.9}$, where $t$ is the current epoch and $T$ is the total number of epochs.

\subsection{Semantic segmentation evaluation details} \label{subsec:semseg_details}
Models are trained (without freezing any layers) using stochastic gradient descent with an initial learning rate of $0.02$, which is decayed by a factor of $0.9$ every 500 iterations, and a constant momentum rate of $0.9$. All models are trained with a batch size of two for 150k iterations in total.

\subsection{Computing correlations} \label{subsec:correlation_details}
\begin{figure*}[t]
    \centering
    \includegraphics[width=0.95\textwidth]{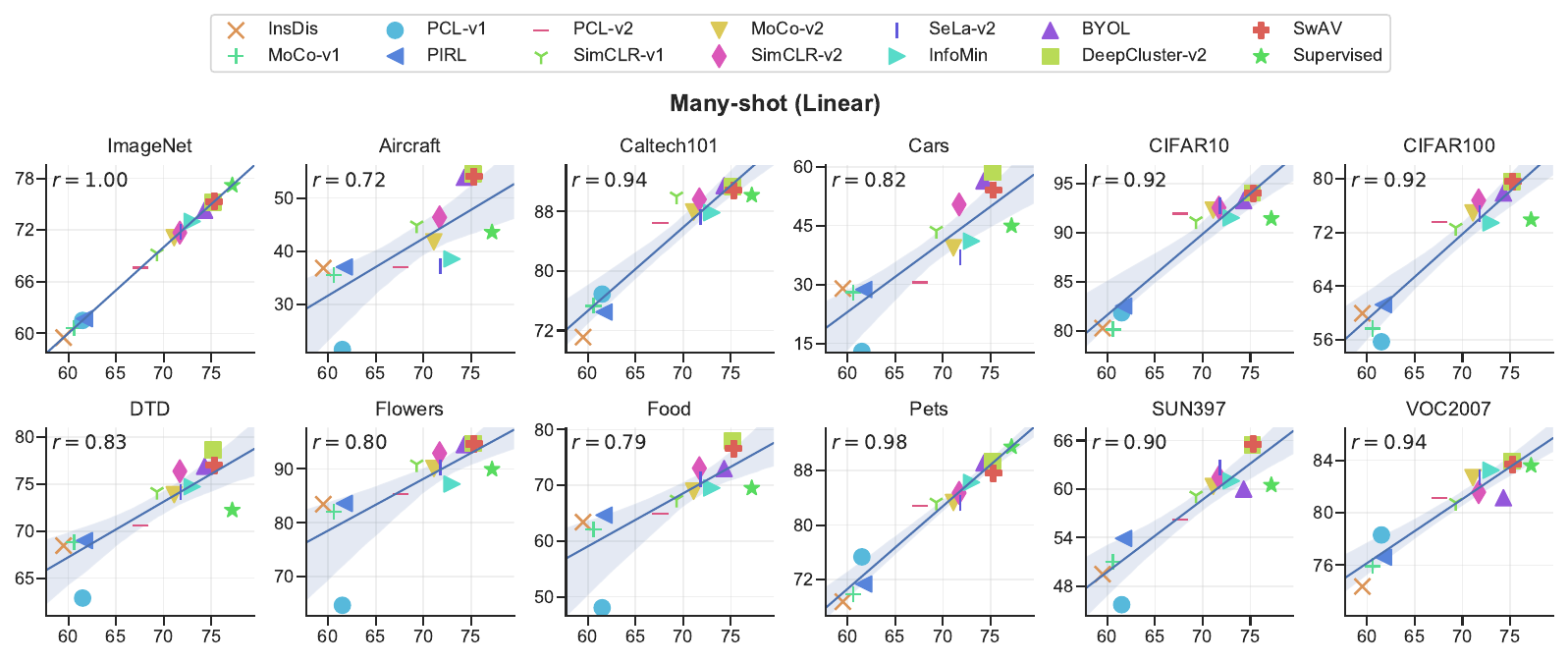}
    \includegraphics[width=0.95\textwidth]{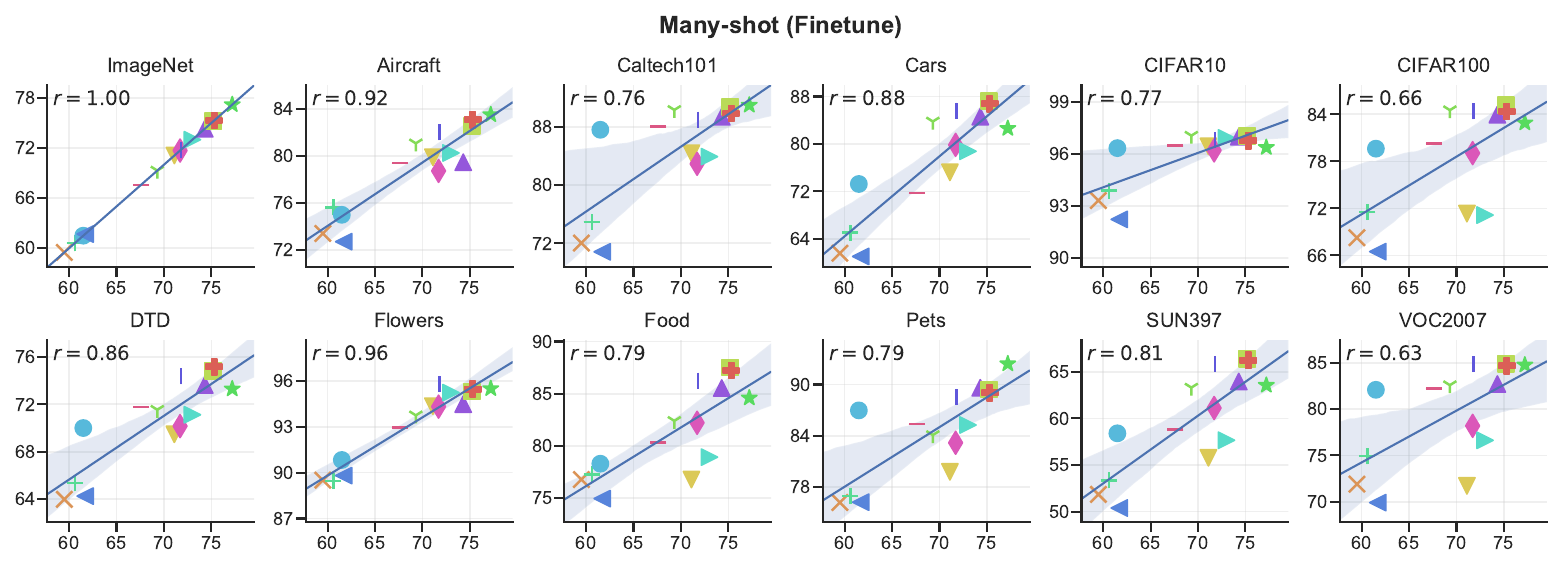}
    \includegraphics[width=0.76\textwidth]{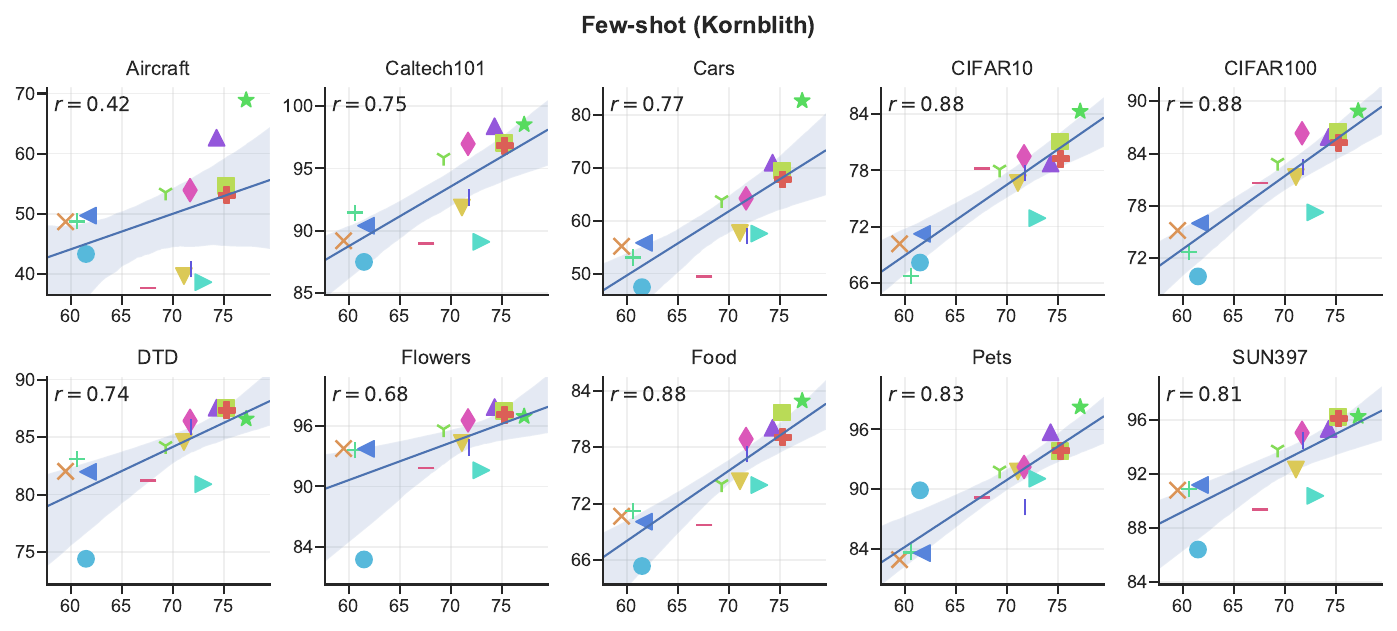}
    \includegraphics[width=0.62\textwidth]{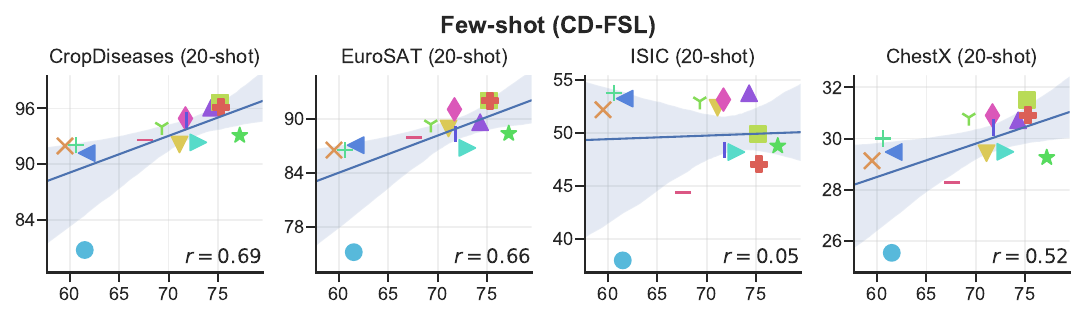}
    \caption{Individual plots of transfer correlations between ImageNet accuracy on the x-axis and target performance on the y-axis.}
    \label{fig:transfer_grid}
\end{figure*}

\begin{figure*}[t]
    \centering
    \includegraphics[width=0.95\textwidth]{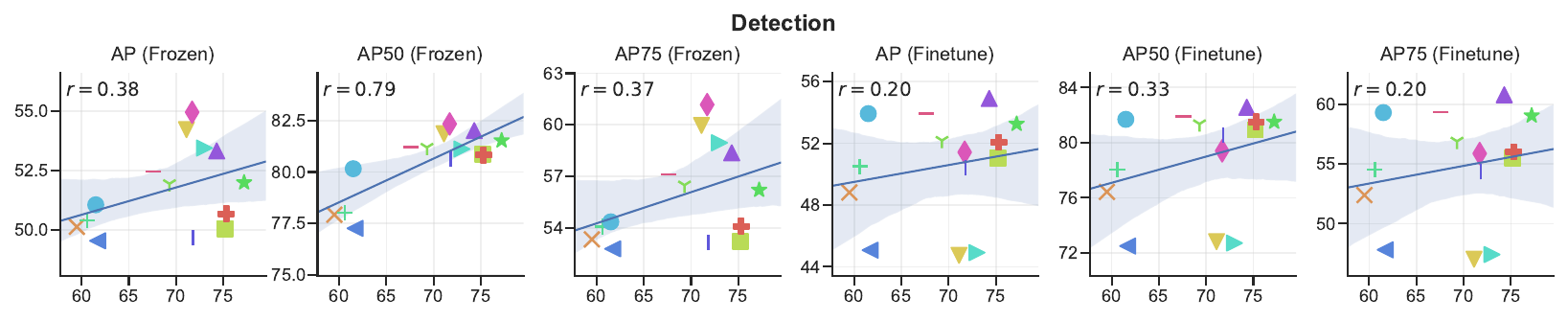}
    \includegraphics[width=0.78\textwidth]{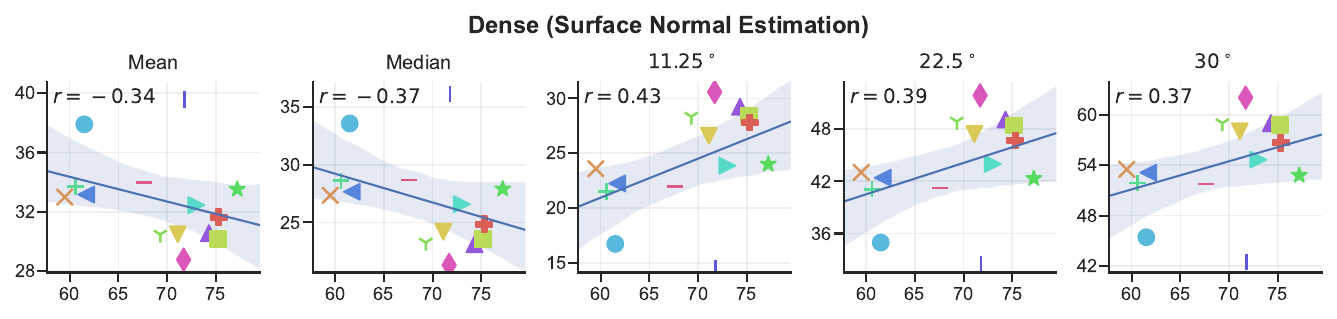}
    \includegraphics[width=0.33\textwidth]{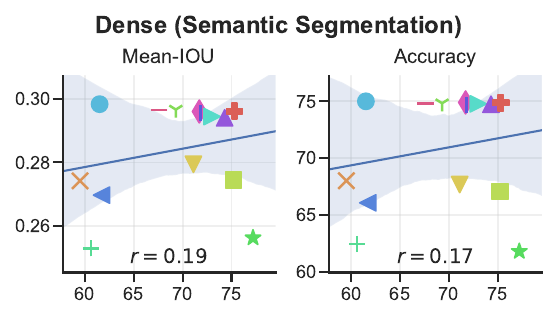}
    \caption{Individual plots of transfer correlations between ImageNet accuracy on the x-axis and target performance on the y-axis.}
    \label{fig:transfer_grid_2}
\end{figure*}

At many points in this work we analyse the statistical relationships between different results. This includes the correlation coefficients in Figs.~\ref{fig:transfer_full}, \ref{fig:correlations}, \ref{fig:calibration}, \ref{fig:transfer_grid} and \ref{fig:transfer_grid_2}, those reported in the text and more summarised in Table~\ref{tab:qualitative_results}. In order to capture the fact that an absolute increase of 1\% in accuracy has varying significance depending on if, e.g., the accuracy goes from 50\% to 51\% or if it goes from 98\% to 99\%, we apply a logit-transformation to any metric that is bounded in the range $0$ to $1$.

All correlations computed against ImageNet performance use the logit-transformed ImageNet top-1 accuracy. Additionally, we logit-transform all recognition accuracies, AP metrics from detection, $11.25^\circ$, $22.5^\circ$ and $30^\circ$ in surface normal estimation, and both mean-IOU and accuracy in semantic segmentation. The only metrics not transformed in this way are the Mean and Median errors in surface normal estimation. We negate these two error metrics before computing correlations in Fig.~\ref{fig:correlations} so reading the figure is easier.

For correlations in Fig.~\ref{fig:transfer_full}, we average the logit-transformed accuracies across datasets in all many-shot and few-shot settings to produce a single correlation coefficient for each setting. For both detection settings we report the correlation of the logit-transformed AP50 metric and for the two dense settings we report correlations of the logit-transformed $11.25^\circ$ and mean-IOU metrics.

For calibration (Fig.~\ref{fig:calibration}), perceptual similarity and attentive diffusion (Table~\ref{tab:qualitative_results}), we similarly use logit-transformed values when computing correlations. For the red, green and blue colour channel errors in our image reconstruction, we report correlations of their raw values.

\subsection{Image reconstruction by feature inversion} \label{subsec:recon_details}
To see what information is retained by the models, we evaluate how well an image can be reconstructed from an extracted feature. We follow the deep image prior \cite{Ulyanov2018DeepPrior} protocol of feature inversion. Given an image $I$, we first extract its feature vector $f(I)$ by passing it through the pre-trained model backbone $f$. Next, we initialise a reconstruction network $g_\theta$, parameterised by $\theta$, which maps from a fixed noise input $z$ to an image $g_\theta(z)$. The reconstruction network is then trained to output an image which, when passed through our pre-trained backbone, produces a feature close to that of $I$. The optimisation problem is:
\setlength{\belowdisplayskip}{2pt} \setlength{\belowdisplayshortskip}{2pt}
\setlength{\abovedisplayskip}{4pt} \setlength{\abovedisplayshortskip}{4pt}
\begin{equation}
    \argmin_\theta f(g_\theta(z)) - f(I).
\end{equation}

We extract the features from our pre-trained backbone from the 4th residual block, giving a vector size of $2048 \times 7 \times 7$. The reconstruction network is trained for 3000 iterations using the Adam optimiser with a learning rate of 0.001. The architecture of the reconstruction network is the same as in the original deep image prior paper \cite{Ulyanov2018DeepPrior} and the study in \cite{Zhao2020WhatLearning}.

\subsection{Computing the saliency maps} \label{subsec:saliency_details}
We use an occlusion mask of $10 \times 10$ pixels and pass it over images resized to $242 \times 242$ which we then crop to $224 \times 224$ to ensure all pixels are occluded the same number of times. The attention values are computed as the root relative squared error (RRSE) of the original features and the occluded features, averaged over all times a pixel is occluded ($10^2$). The RRSE ensures that the distances are invariant to the scale of the original features.

\begin{figure*}[t]
    \centering
    \includegraphics[width=0.4\textwidth]{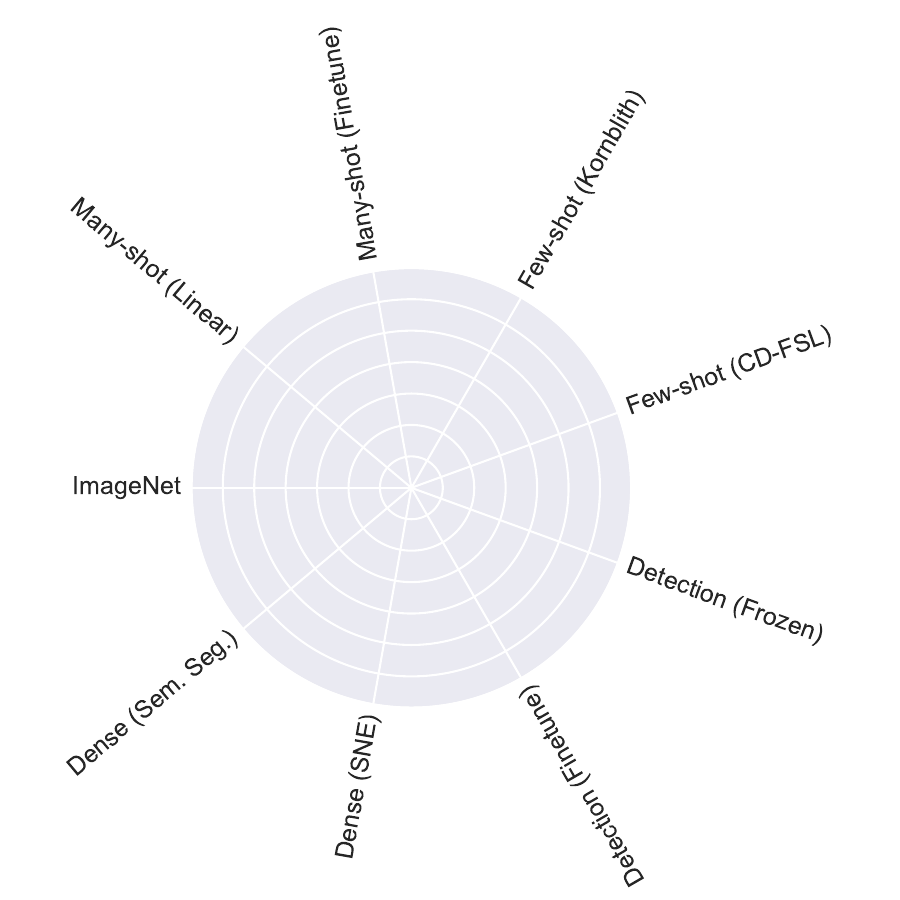}
    \includegraphics[width=\textwidth]{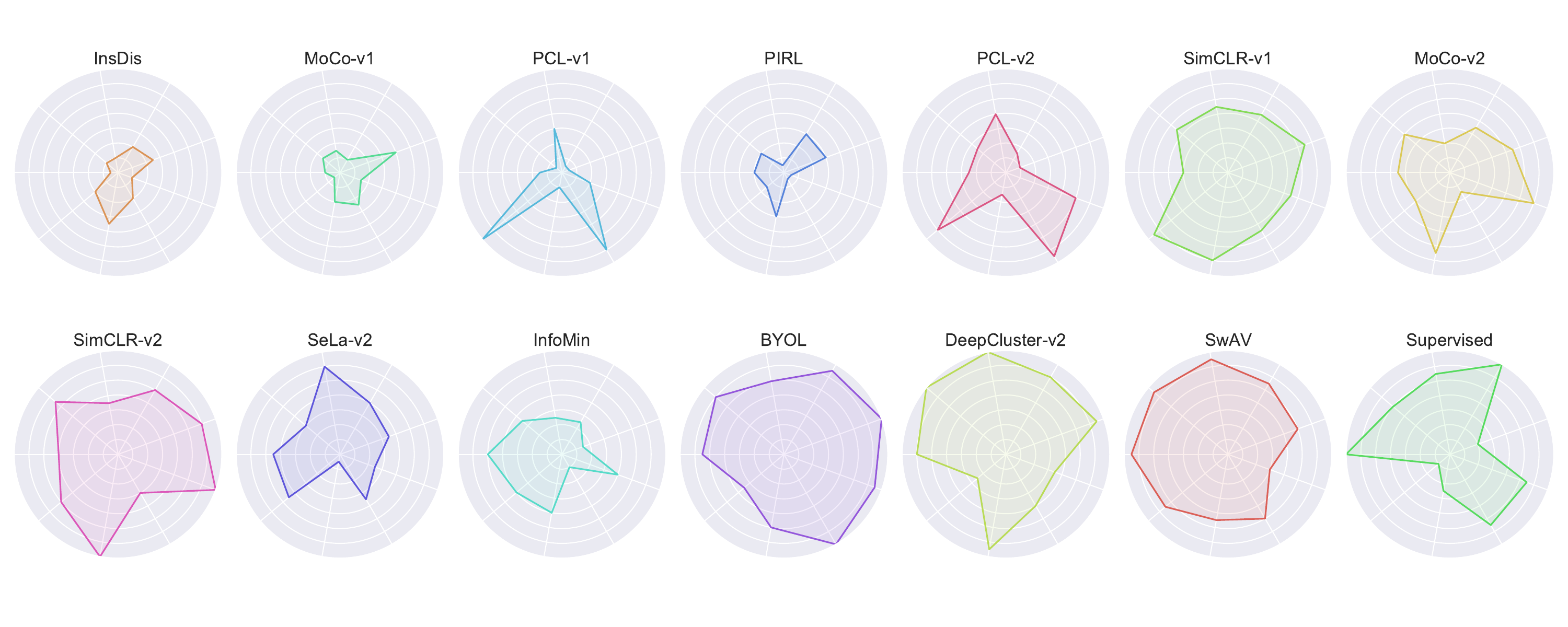}
    \caption{Radar charts of model performance on ImageNet and our eight different evaluation settings. In each setting we compute the rankings of the models (from averaged performance where there are multiple datasets). In each plot above, a higher rank (better performance) places the line closer to the outer edge of the circle. A larger total area roughly corresponds to better performance across a wide range of transfer settings. The rankings are based on average accuracy in the many-shot and few-shot settings, AP50 for frozen and finetuned detection, mean error for surface normal estimation and mean IOU for semantic segmentation.}
    \label{fig:radar_charts}
\end{figure*}

\begin{figure*}[t]
    \centering
    \includegraphics[width=\textwidth]{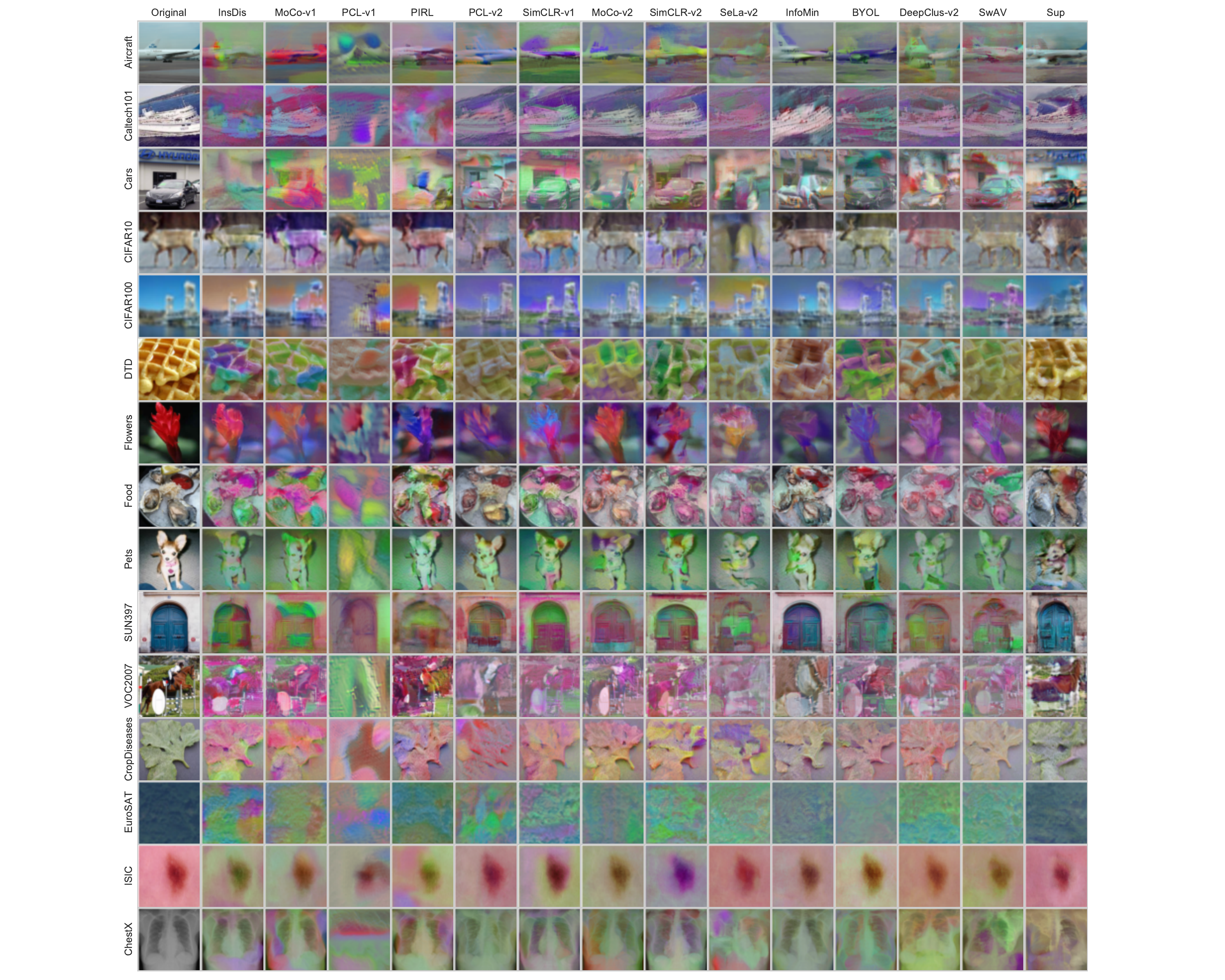}
    \caption{Deep image prior reconstructions on one image for each of 15 datasets.}
    \label{fig:recon_grid}
\end{figure*}

\begin{figure*}[t]
    \centering
    \includegraphics[width=\textwidth]{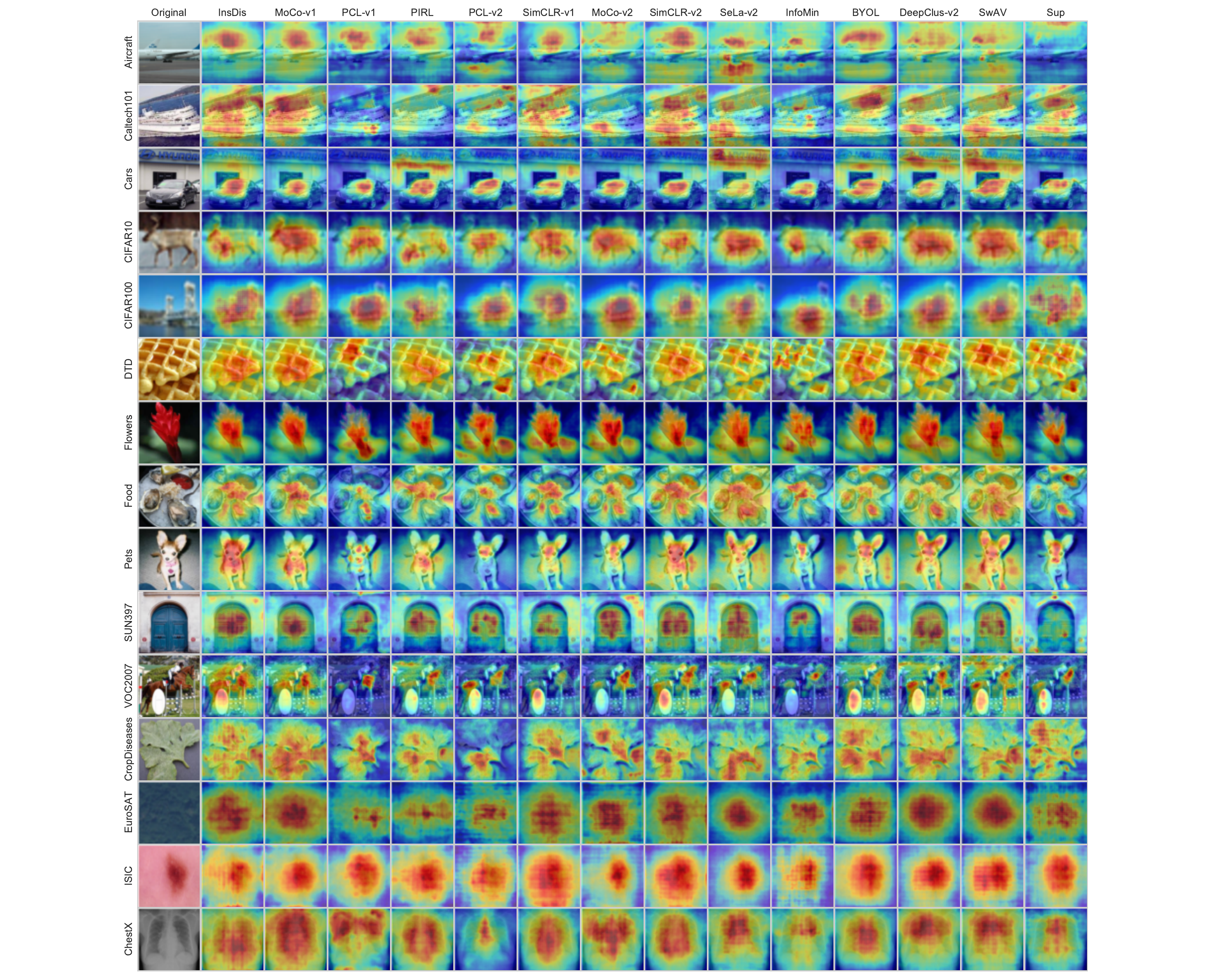}
    \caption{Saliency maps for all models on one image for each of 15 datasets.}
    \label{fig:attention_grid}
\end{figure*}